\documentclass[preprint,1p,authoryear]{elsarticle}

\usepackage{amssymb}
\usepackage{dirtytalk}
\usepackage{graphicx}
\usepackage{subfig}
\usepackage{times}
\usepackage{epsfig}
\usepackage{capt-of}
\usepackage{latexsym}
\usepackage{amsmath}
\usepackage{setspace}
\usepackage{epsfig}
\usepackage{multirow}
\usepackage{url}
\usepackage{fixltx2e}
\usepackage{mathtools}
\usepackage{color}
\usepackage{todonotes}
\usepackage{makecell}

\usepackage{etoolbox}
\usepackage{lineno}
\usepackage{hyperref}

\modulolinenumbers[5]

\patchcmd{\pprintMaketitle}
 {\hrule}
 {\clearpage\hrule}
 {}{}
\appto\endfrontmatter{\clearpage}

\begin{document}

\newcommand{\longedge}[3]{{#1}\buildrel{#2}\over\longrightarrow{#3}}
\begin{frontmatter}



\author[addr1]{Tomer Wullach\corref{cor1}}
\ead{tomerwullach@gmail.com}
\author[addr2]{Amir Adler}
\ead{adleram@braude.ac.il}
\author[addr1]{Einat Minkov}
\ead{einatm@is.haifa.ac.il}
\address[addr1]{Department of Information Systems, University of Haifa}
\address[addr2]{Department of Electrical Engineering, Braude College of Engineering, Karmiel, Israel, and\\ McGovern Institute for Brain Research, MIT, Cambridge, MA, USA.}
\cortext[cor1]{Corresponding author}

\title{Character-level HyperNetworks for Hate Speech Detection}


\author{} 

\address{}
\clearpage
\begin{abstract}
The massive spread of \textit{hate speech}, hateful content targeted at specific subpopulations, is a problem of critical social importance. Automated methods of hate speech detection typically employ state-of-the-art deep learning (DL)-based text classifiers--large pretrained neural language models of over 100 million parameters, adapting these models to the task of hate speech detection using relevant labeled datasets. Unfortunately, there are only a few public labeled datasets of limited size that are available for this purpose. We make several contributions with high potential for advancing this state of affairs. We present HyperNetworks for hate speech detection, a special class of DL networks whose weights are regulated by a small-scale auxiliary network. These architectures operate at character-level, as opposed to \textcolor{black}{word or subword-level}, and are several orders of magnitude smaller compared to the popular DL classifiers. We further show that training hate detection classifiers using additional large amounts of automatically generated examples is beneficial in general, yet this practice especially boosts the performance of the proposed HyperNetworks. \textcolor{black}{We report the results of extensive experiments, assessing the performance of multiple neural architectures on hate detection using five public datasets. The assessed methods include the pretrained language models of BERT, RoBERTa, ALBERT, MobileBERT and CharBERT, a variant of BERT that incorporates character alongside subword embeddings. In addition to the traditional setup of within-dataset evaluation, we perform cross-dataset evaluation experiments, testing the generalization of the various models in conditions of data shift. Our results show that the proposed HyperNetworks achieve performance that is competitive, and better in some cases, than these pretrained language models, while being smaller by orders of magnitude.}
\end{abstract}

\begin{keyword}
Hate speech detection\sep Neural Networks\sep Text Generation
\end{keyword}
\end{frontmatter}


\section{Introduction}

Social media platforms are strongly  criticised nowadays for enabling the spread of hate speech--the expression of hate or encouraging violence based on race, nationality, religion, and sexual orientation, among others. While social media providers take some measures for fighting hate speech, such as the sampling and screening of suspicious content by dedicated systems and workforce, these measures fail to identify and mitigate the manifestation of hate speech at scale. To this end, automatic methods for detecting hate speech are being developed, where many challenges are yet to be addressed. 

\textcolor{black}{In accordance with the advances of neural computing methods for text processing, recently proposed methods of automatic hate detection employ high-performing large neural network architectures, mainly, the transformer-based BERT~\citep{devlin2018bert} and its variants, e.g.,~\citep{tranEMNLP20,sarkarEMNLP21,miok2021ban}. These models are pretrained as contextual encoders of token and sentence meaning using vast amounts of general unannotated texts. In classification, the underlying meaning of a given text is encoded by these models onto a low-dimension vector space, which is then mapped to specified target categories given labeled examples in another learning step named as {\it fine-tuning}.}

One key challenge of learning automatic models of hate detection is the paucity and lack of diversity of labeled hate speech examples. Since the vast majority of social media content is not offensive or hateful, a massive amount of data needs to be screened in order to collect a sufficient and representative sample of hate speech examples. Researchers therefore revert to collecting hate speech samples that contain specified terms, or focus on the content posted by accounts that are known to be hateful. Quality labeling demands reaching an inter-coder agreement regarding the correct label for every sample, making the data annotation process complicated, time consuming and expensive. For these reasons, hate speech datasets are in most cases small, topically-biased, and imbalanced, including relatively few hate speech examples~\citep{wiegand2019detection,wullach2020towards}. \textcolor{black}{Several recent works investigated methods for automatically expanding the existing datasets by means of data augmentation~\citep{fengACL21}, for example by means of back translation and paraphrasing~\citep{rizos2019augment,beddiarJO21}}. In this work, we follow the approach by Wullach et al.~\citeyearpar{wullach2020towards}, augmenting the available labeled datasets using synthetic hate and non-hate examples generated by  GPT-2~\citep{radford2019language}, a pretrained generative language model which we fine-tune to synthesize text sequences that are similar yet diversify the available relevant examples. The datasets that we synthesized in this work are balanced and an order of magnitude larger than any previously explored hate speech datasets, accumulating in total to 10M generated text sequences. 

Another main challenge of identifying hate speech automatically is that the textual content posted on social media platforms (e.g., Twitter) in general, and hateful expressions in particular, \iffalse Hate speech data obtained from \fi are often noisy and diverse language-wise, as users tend to quickly adopt new terms and use creative and personalized language styles. \iffalse as hate speech has many flavors. This property imposes another challenge in training a \fi State-of-the-art  deep-learning hate speech detectors exhibit vocabulary limitations, using a memory expensive lookup table which maps each text \textcolor{black}{token} to a high-dimensional embedded vector~\citep{mikolov2013distributed}. These models are therefore challenged by the rich diversity and irregularity of social media language~\citep{achilles20}. 

Finally, a challenge of practical importance is that state-of-the-art deep learning networks are extremely large, reaching up to hundreds of millions, or billions of parameters. \textcolor{black}{The large size of these models and their high latency prevent their deployment at large scale, or in conditions that are resource-limited, such as mobile edge computing~\citep{7879258}.} Ideally, hate detection would be performed in real time at the end device, allowing to alert the user prior to posting hateful content, yet high-performing compact hate detection models are required to achieve this goal~\citep{tranEMNLP20,mitraICSC21}. In order to alleviate the associated deployment costs, researchers have recently proposed to distill the large pretrained transformer-based models into smaller neural architectures, typically reducing the original model size by up to one magnitude of order, e.g.,~\citep{sanh2019distilbert,tinyBERT,mobileBert}.

In this work, we introduce new learning architectures of hate speech detection that are based on HyperNetworks, a special class of deep learning networks, which utilize weight sharing across layers~\citep{ha2016hypernetworks}. The proposed architectures process the text at character-level--as opposed to \textcolor{black}{token}-level. This results in a significant reduction in the total number of learnable parameters, introducing hate detection classifiers that are extremely efficient and compact. 

\textcolor{black}{We assess the proposed models alongside several deep learning networks, including the popular large transformer-based BERT and RoBERTA~\citep{roberta19} architectures, which model text at wordpiece level. We further consider in our experiments the smaller transformer-based models of  ALBERT~\citep{albert} and MobileBERT~\citep{mobileBert}, a recently introduced distilled variant of BERT, as well as the recently proposed model of CharBERT~\citep{charbert}, which introduces character embeddings alongside the wordpiece information into BERT. In addition, we experiment with CNN-GRU~\citep{zhang2018detecting}, a smaller neural network model that has previously shown strong results on the task of hate detection, modeling text at word level. Our empirical evaluation applies to five public datasets labeled for hate speech detection. We train and assess the models using examples drawn from from a single dataset, as well as report cross-dataset evaluation results, where the models are trained and tested using examples drawn from different datasets. The latter scenario is more challenging, and more realistic in assuming a possible shift in the underlying data distribution. In order to improve generalization, we train the models in both setups using increasing amounts of training examples, leveraging our large-scale resource of synthetic examples.}  

\textcolor{black}{Our findings extend previous research~\citep{wullachEMNLP21}, showing that all of the above networks benefit from data augmentation, especially in data shift conditions, and mainly due to consistent improvements in recall. We find that while the HyperNetwork models are inferior to the larger models given limited amounts of labeled data, training the HyperNetworks using increasing amounts of generated labeled data, up to 1-2M text sequences, consistently improves and achieves competitive and in some cases even better performance than the popular deep learning methods, while requiring only a fraction of the number of parameters.}

Based on these results, we believe that the proposed networks, which include only tens of thousands of parameters, pave the way to \textcolor{black}{mobile edge computing~\citep{7879258}, allowing} automatic hate speech deployment in end-user devices with limited computation resources. \iffalse including smartphones and tablets.\fi

The main contributions of this paper are three-fold:
(1) the paper presents for the first time the utilization of HyperNetworks for hate speech detection. The proposed networks operate at character-level, and have an exceptionally low number of parameters.
(2) by increasing significantly the training data set size via text generation, the proposed solutions are demonstrated to achieve competitive, or better performance in some cases, than state-of-the-art deep learning models, which are orders of magnitude larger than the proposed solutions. (3) the paper is accompanied by a new hate speech corpus that is the largest ever created (10M sequences).\footnote{The created dataset will be shared for research purposes upon request to the authors.} This new corpus results from data generation using a state-of-the-art deep generative language model, fine-tuned to approximate the distributions of five hate speech public datasets. \textcolor{black}{Our results indicate that all hate detection models benefit greatly from training using this resource.}

\textcolor{black}{The rest of the paper is organized as follows: Section 2 reviews  hate speech detection methods and datasets, Section 3 presents the proposed HyperNetworks-based  approach, Section 4 provides a detailed performance evaluation, and Section 5 concludes the paper.}

\section{Background: Hate Speech Methods and Data Resources}

Research has proposed several approaches for hate speech detection over the past years, recently employing deep learning approaches. Deep learning classification models are known to perform and generalize well when a large, diverse, and high-quality data resource is available for training. In the lack of such a resource, previous related works constructed datasets for this purpose, which were manually labeled, and are therefore strictly limited in size. 

In this Section, we describe several popular public labeled datasets that have been used by researchers for training and evaluating hate speech detectors. Following previous works~\citep{wullach2020towards,wullachEMNLP21}, we exploit a large-scale corpus of generated hate and non-hate sequences that extend these datasets for training the various hate classification methods included in our experiments.

This Section further includes a general overview of existing approaches of hate speech detection, including a detailed description of the hate speech classification methods that we apply in this work. Later, we review related literature concerning character-based convolution neural networks and parameter generation via HyperNetworks, as we utilize this methodology in proposing compact and efficient character-based hate detectors.

\subsection{Hate speech datasets}
\label{ssec:datasets}

The typical process that has been traditionally applied by researchers and practitioners for constructing labeled datasets in general, and of hate speech in particular, involves the identification of authentic hate speech sentences, as well as counter (non-hate) examples. Twitter is often targeted as a source of relevant data; this (and other) public social media platform allows users to share their thoughts and interact with each other freely, making it a fertile ground for expressing all kinds of agendas, some of which may be racist or hateful. The initial retrieval of hate speech examples from Twitter is based on keyword matching, specifying terms that are strongly associated with hate.\footnote{e.g., \url{https://www.hatebase.org/}} Once candidate tweets are collected, they are assessed and labeled by human annotators into pre-specified categories. 
The manual annotation of the examples is intended to result in high-quality ground truth labeled datasets. Yet, manual annotation is costly~\citep{modhaESWA20}, and accordingly, the available datasets each include only a few thousands of labeled examples. Due to their size limit, and  biases involved in the dataset collection process, e.g., keyword selection and labeling guidelines, these datasets are under-representative of the numerous forms and shapes in which hate speech may be manifested~\citep{wiegand2019detection}.\footnote{Another caveat of referring to authentic content within a dataset concerns the discontinued availability of the collected texts by the relevant provider over time. For example, tweets must be stored by their identifier number, where access to the tweet's content may be defined, impairing the dataset.} 

Table~\ref{tab:datasets} details the statistics of five popular public datasets of hate speech, which we experiment with in this work. All of the datasets were manually curated and are of modest size, including a few thousands of labeled examples, out of which a minority contain hate speech. 

Let us describe the individual datasets listed in Table~\ref{tab:datasets} in more detail. The dataset due to Davidson et al.~\citeyearpar{davidson2017automated} (\textbf{DV}) includes tweets labeled by CrowdFlower\footnote{https://www.welcome.ai/crowdflower} workers into three categories: {\it hate speech}, {\it offensive}, or {\it neither}. For the purposes of this work, we only consider the examples of the first and latter categories.  Waseem and Hovy~\citeyearpar{waseem2016hateful} created another dataset (\textbf{WS}), considering tweets of accounts which frequently used slurs and terms related to religious, sexual, gender and ethnic minorities; those tweets were manually labeled into the categories of {\it racism}, {\it sexism} or {\it neither}. Again, as we focus on hate speech, and strive at compatible labels across datasets, we only consider the first and latter categories as examples of hate and non-hate, respectively. Another dataset was constructed by SemEval conference organizers~\citep{basile2019semeval} for the purpose of promoting hate detection (\textbf{SE}). They considered the historical posts of identified hateful Twitter users, narrowed down to tweets that included hateful terms, and had examples labeled by CrowdFlower workers. Many of the tweets labeled as hateful in this dataset target women and immigrants. Founta et al.~\citeyearpar{founta2018large} (\textbf{FN}) performed iterative sampling and exploration while having tweets annotated using crowdsourcing. Their resulting dataset is relatively large ($\sim$80K examples), and distinguishes between multiple flavors of offensive speech, namely {\it offensive}, {\it abusive}, {\it hateful}, {\it aggressive}, {\it cyber bullying}, {\it spam} and {\it none}. In order to maintain our focus on hate speech, we consider the labeled examples associated with the {\it hateful} and {\it none} categories. Finally, the dataset due to de-Gibert et  al.~\citeyearpar{de-gibert18} (\textbf{WH}) was extracted from the extremist StormFront Internet forum,\footnote{https://en.wikipedia.org/wiki/Stormfront\_(website)}. This dataset aims to gauge hate in context, considering also cases where a sentence does not qualify as hate speech on its own, but is interpreted as hateful within a larger context comprised of several sentences. 

We leveraged these datasets for (a) training and evaluating several types of hate speech detectors, (b) fine-tuning a pre-trained generative model and automatically producing a large number of additional similar hate speech sequences for training purposes. In our experiments, we split the available examples into fixed train (80\%) and test (20\%) sets, while maintaining similar class proportions. Only the train examples are used in the sequence generation process. Additional details about these datasets, as well as examples of the tweets generated per dataset, are available in Wullach {\it et al.}~\citeyearpar{wullach2020towards}. 

\textcolor{black}{While our experiments concern hate speech detection in English, our approach is language-independent. Being semi-supervised, it requires relevant labeled examples in the target language, as well as a language-specific model decoder such as GPT, for expanding the manually labeled datasets by generating additional examples synthetically. Such resources in non-English languages are currently scarce but are being developed~\citep{multilingual,deVriesACL21}.}

\begin{table}[t]
\small
\centering
\caption{Publicly available hate speech datasets: size statistics}
\label{tab:datasets}
\setlength\tabcolsep{0pt}
\begin{tabular*}{0.7\columnwidth}{@{\extracolsep{\fill}} lrc}
      Dataset & Size & Hate class \%\\\hline\hline
%\midrule

{~\cite{davidson2017automated}  (\textbf{DV})} &  {6K} & {24\%}\\
{~\cite{founta2018large} (\textbf{FN})} &  {53K} & {11\%}\\ 
{~\cite{waseem2016hateful} (\textbf{WS})} & {13K} & {15\%}\\
{~StormFront~\citep{de-gibert18} (\textbf{WH})} &  {9.6K} & {11\%}\\
{~SemEval2019~\citep{basile2019semeval} (\textbf{SE})} & {10K} & {40\%}\\
\hline
%\bottomrule
\end{tabular*}
\end{table}

\subsection{Hate Speech Detection}

Deep learning (DL) can effectively exploit large-scale data, learning latent representations using multi-layered neural network architectures. Various modern DL architectures of text classification consist of a word embedding layer \textcolor{black}{(a dimensionality reduction (\cite{ding2021dimensionality}) technique that is commonly utilized in DL-based natural language processing models)} that is intended to capture generalized semantic meaning of words, mapping each word in the input sentence into a vector of low-dimension~\citep{mikolov2013distributed}. The following layers learn relevant latent feature representations, where the processed information is fed into a classification layer that predicts the label of the input sentence.

For example, previously, Founta et al.~\citeyearpar{founta2019unified} employed the following DL architecture for detecting hate speech (and other types of offensive texts). They transformed the input words into GloVe word embeddings~\citep{pennington2014glove}. They then used a recurrent layer comprised of Gated Recurrent Unit (GRU) for generating contextual and sequential word representations, having each word processed given the representations of previous words in the input sentence. Following a dropout layer (intended to prevent over-fitting), the final dense layer outputs the probability that the sentence belongs to each of the targets using a softmax activation function. Other works on hate speech detection employed Convolution neural networks (CNN) for modeling contextual information, e.g.,~\citep{badjatiya2017deep}. CNN applies a filter over the input word representations, processing local features per fixed-size word subsequences. Hierarchical processing, which involves aggregation and down-sampling, consolidates the local features into global representations, which are fed into the final layer that predicts the probability distribution over the target classes.

In this work, we consider a popular DL classification architecture due to Zhang et al.~\citeyearpar{zhang2018detecting}, which comprises both convolution and recurrent processing layers. \textcolor{black}{This architecture has been previously shown to yield top-performance results on the task of hate speech classification, and is therefore evaluated as one of the architectures of choice in our experiments, similar to other related studies~\citep{hatexplain}.} In brief, this text classification network consists of a convolution layer applied to the input word embeddings, which is down-sampled using a subsequent max-pooling layer. The following layer is recurrent (GRU), producing hidden state representations per time step. Finally, a global max-pooling layer is applied, and a softmax layer produces a probability distribution over the target classes for the given input. Further implementation details are provided later in this paper.

\subsection{Hate Speech Detection with Pre-trained Language Models}

\textcolor{black}{More recently, larger and deeper language models (LMs), that have been pretrained on massive heterogeneous corpora, were shown to yield state-of-the-art contextual text representations, leading to further improvement in text classification. We focus our attention on the popular transformer-based language encoder of BERT~\citep{devlin2018bert} and its variants, as described below.} 

\paragraph{BERT}

Text classification using pretrained language models like BERT~\citep{devlin2018bert} has been shown to give state-of-the-art performance on a variety of text processing tasks. While BERT generates task-agnostic contextual word embeddings, it can be optimized to target tasks via {\it fine-tuning}~\citep{devlin2018bert}. We follow the common practice of turning BERT into a text classifier by adding a final feed-forward network, which receives as input the embedding of the input text as processed by BERT (the `[CLS]' token embedding, which has been tuned to represent the meaning of the whole input text sequence), and outputs the target class probabilities via a softmax layer. Given labeled training data, the extended network parameters, including the weights of the terminal network, as well as BERT parameters, are fine-tuned jointly to optimize classification performance. 

\paragraph{RoBERTa}

This model applies the same architecture as BERT, but has been trained on ten times more data, including news articles and Web content. Due to this augmentation of training data, and other modifications to the pretraining procedure and cost function, RoBERTa has been shown to outperform BERT on multiple benchmark datasets~\citep{roberta19}.

\paragraph{ALBERT}

The architecture of ALBERT~\citep{albert} was designed as a light variant of the transformer-based BERT model. It enhances BERT in several ways, including a factorization of the embedding parameters and cross-layer parameter sharing; both measures are intended to improve parameter efficiency as well as a form of regularization. ALBERT also replaces the next-sentence-prediction loss that is used in training BERT with sentence-order prediction loss, which focuses on modeling inter-sentence
coherence. As a result, ALBERT has been shown to outperform BERT on several multi-sentence encoding tasks~\citep{albert}. \\

It has been shown that while BERT has been pretrained using vast amounts of textual data, namely, all of Wikipedia and Google Books, the word encodings it produces can benefit from further adaptation to the target domain and task~\citep{gururACL2020}. Indeed, there exist multiple works that aim to adapt pretrained models like BERT to the task of hate detection using additional relevant data. Isaksen and Gamb{\"a}ck~\citeyearpar{isaksen2020} report results of fine-tuning the base and large variants of BERT on each of the \textbf{DV} and \textbf{FN} datasets. In addition to the pretrained version of these methods, they adapted the models by further training them using unlabeled examples from additional hate detection datasets. Overall, they found the performance of BERT-base to be comparable or better than BERT-large, but failed to improve the model results using the unlabeled data. Another work introduced HateBERT~\citep{hatebert}, a model of BERT (base) which was further trained on Reddit comments extracted from communities banned for being offensive, abusive, or hateful. A comparison of BERT and HateBERT on hate detection yielded mixed and non-conclusive results. \textcolor{black}{More recently, Sarkar et al.~\citeyearpar{sarkarEMNLP21} introduced {\it fBERT}, a variant of BERT (base) which they continued to train using large amounts of tweets that were labeled as offensive using semi-supervised classification. Unlike HateBERT, they achieved performance improvements on several experimental datasets.}  

\textcolor{black}{In our experiments, we fine-tune and evaluate BERT, RoBERTa and ALBERT as hate detection classifiers on multiple datasets. Similar to other works~\citep{wiegand2019detection,mozafariPLOS2020,wullachEMNLP21}, we consider both within- and cross-dataset evaluation setups, testing the models generalization when trained and tested on examples drawn from different data distributions. Here and elsewhere~\citep{wullachEMNLP21}, we show that augmenting the available training datasets with additional generated examples results in substantial performance gains and improved generalization.}

\iffalse We propose an approach for detecting hate speech using character-based models. Previous works demonstrated the advantages of utilizing a character-based models, achieving competitive performance by utilizing character-based for detecting hate speech. Such models are composed of a significantly lower number of parameters, making them less complex and better fitted on memory-limited devices. Moreover, character-based models are less sensitive to misspelled and out-of-vocabulary terms compared to models that use a word (or sub-word) vocabulary, an valuable attribute when dealing with data drawn from Tweeter. We also experiment with "Weight Generating Networks"~\citep{ha2016hypernetworks, choi2019adaptive}, allowing the model to share weights across layers and generate model weights that are conditioned on the input.\fi

\textcolor{black}{Admittedly, while transformer-based pretrained language models yield state-of-the-art performance, they are characterised with large model sizes and high latency. Concretely, here we apply the base configurations of BERT and RoBERTa, which both include 110 million parameters. Ongoing efforts aim to design smaller versions of these models with faster inference times by means of distillation. The distilled models of   DistilBERT~\citep{sanh2019distilbert} and TinyBERT~\citep{tinyBERT} have $\sim$40\% and 7.5x less parameters than BERT base, respectively. Here, we experiment with ALBERT, a light and competitive variant of BERT that has 17 million parameters, i.e., $\sim$9x less parameters than BERT base. We further experiment with MobileBERT~\citep{mobileBert}, another recently introduced model that distills the pretrained BERT into a deep and thin model that includes 24 million parameters, being 4.3× smaller compared with BERT base.}

\textcolor{black}{As detailed in Table~\ref{tab:params}, the character-based networks that are proposed and evaluated in this work are smaller by orders of magnitude compared with the transformer-based models, as well as compared with CNN-GRU. Yet, we show that these compact models, when trained using large amounts of relevant data, can achieve competitive levels of performance.}

\subsection{Character-based models}

\textcolor{black}{Character-level CNNs have been previously shown to be successful on several natural language processing (NLP) tasks, e.g.,~\citep{kim2015character, zhang2015character,mehdad2016characters,charSemEval}. Mehdad and Tetreault~\citeyearpar{mehdad2016characters} considered the task of abusive language detection, and showed that light-weight and simple character-based approaches might be superior to token-based modeling using adequate methods on this task; specifically, they experimented with recurrent neural network architectures.} Compared with the character-level deep learning networks proposed previously~\citep{zhang2015character}, our solution is more shallow and compact, as it incorporates HyperNetworks~\citep{ha2016hypernetworks}. The HyperNetwork architecture utilizes a relaxed-form of weight sharing across the network layers, enabling adaptive tuning of the network weights according to specific input text sequences. It therefore promotes generalization, while further reducing the parameter space. 

The popular deep learning architectures that we consider in this work model language at word-level (CNN-GRU) or sub-word level (BERT and its variants). \textcolor{black}{Sub-word lexicons are typically generated by applying a learning algorithm over the training set, aiming to find the most likely vocabulary representing the training data. A prominent subword representation used by modern language models is WordPiece, which assembles a subword lexicon by applying an iterative token merging procedure~\citep{kudoACL18}.} \iffalse and SentencePiece, which treats the text as a stream of characters including the spaces between characters~\citep{kudo2018sentencepiece}}.\fi

An important advantage of character-level as opposed to word or sub-word language processing is its flexibility in handling unknown out-of-vocabulary terms, morphological inflections, and noisy word variants that are prevalent on social media. Conversely, it has been shown that the BERT model is highly sensitive to noise in the data, such as spelling mistakes and word variations~\citep{achilles20}. \textcolor{black}{Previously, several deep contextual language modeling architectures, such as ELMO~\citep{elmo}, incorporated character-level information. The recently proposed model of {\it CharBERT}~\citep{charbert} enhances the BERT and RoBERTa models by fusing the representations of characters and subwords, and applies a new pre-training task named NLM (Noisy LM) for unsupervised character representation learning. Overall, CharBERT adds 5M parameters to BERT or RoBERTa, modeling a character channel in addition to the token channel. We experiment with the CharBERT architecture in this work. Unlike CharBERT, the proposed HyperNetwork architectures model the text merely as character sequences, and are orders of magnitude smaller compared with the alternative approaches.}

While character-level text processing may place less emphasis on encoding high-level relationships between words~\citep{zhang2018detecting}, this approach is substantially more compact, and requires modest memory resources in comparison to word-level deep networks.

We describe and motivate the use of several variants of character-level HyperNetworks for hate detection (Section~\ref{sec:charhyper}). 
We further show that such light-weight modeling yields high performance when provided with sufficient amounts of task-specific labeled data, that allows to learn relevant semantic and grammatical phenomena, as well as reduce generalization error~\citep{zhang2015character}. Our approach of generating large amounts of hate- and non-hate text sequences for training these models is described in Section~\ref{sec:gen}.

\section{The Proposed Approach: Character-level HyperNetworks}
\label{sec:approach}

\begin{figure*}[t]
\centering
\includegraphics[trim={0.0cm 13cm 0cm 1cm},clip,scale=0.5]{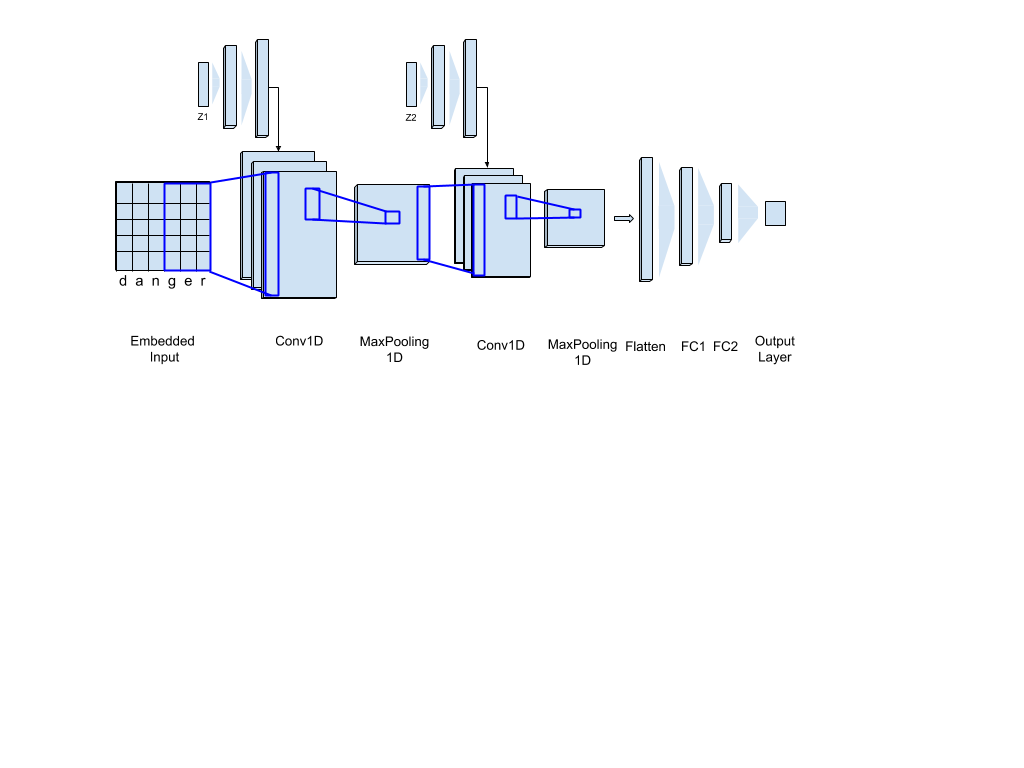}
\caption{The proposed HyperHate-Static architecture: The main network (lower part) computes the posterior probability that an input text sequence is hateful. The weights of the two 1D-convolutional layers are generated by the auxiliary network (upper part). The figure shows for illustration purposes a separate auxiliary network per convolutional layer, however, only a single auxiliary network is implemented and the weights for each convolutional layer are determined by the corresponding layer embedding vector $\mathbf{z}_j$ for $j=1,2$, respectively. The output of the auxiliary network is utilized as weights of each 1D-Convolutional layer, according to  (\ref{eqref:static}).}
\label{fig:HyperHate}
\end{figure*}

\subsection{Character-level HyperNetworks for Hate Speech Detection}
\label{sec:charhyper}

HyperNetworks generally include two sub-networks: a main network--the hate speech detector in our task, and an auxiliary network that generates the weights of the main network.\footnote{In some publications, the auxiliary network is referred to as the \textit{hypernetwork}, yet, in order to avoid ambiguity we refer only to the complete architecture, composed of the main and auxiliary networks, as the \textit{hypernetwork}.} Formally, let $F(\mathbf{x};\mathbf{w}): X\times W \rightarrow Y$ denote the main network, where $\mathbf{x}$ is the input text to be classified, $\mathbf{w}$ are the parameters of the main network and $\mathbf{y}$ is the class label (capital letters represent the space of each variable). The auxiliary network is defined as $G(\mathbf{z};\theta): Z\times \Theta \rightarrow W$, where $\mathbf{z}$ is the auxiliary network input and $\theta$ are the parameters of the auxiliary network. Thus, the auxiliary network computes the weights of the main network, and provides a regularization mechanism over the main network's weights~\citep{ha2016hypernetworks}. The weight computation mechanism can be either independent of the input text during inference, resulting in a \textit{static} hypernetwork, or input-dependent (i.e. adaptive) during inference, resulting in a  \textit{dynamic} hypernetwork.

Next, we first describe the details of the main network, and then present two hate speech detection solutions: (i) a static hypernetwork; and (ii)  a dynamic  hypernetwork. 

\paragraph{The main network} 
Our main network is a character-level CNN, which computes the posterior probability that the input text sequence is hateful. This architecture is inspired by the CNNs proposed in~\citep{zhang2015character}, however, our model is more shallow, comprising of two convolutional layers, compared to six in their work.\\

The text at the network input is represented using an alphabet of 69 characters, comprised of lower-case English letters, digits and other characters, as detailed in Table \ref{tab:alphabet}. We conducted a preliminary study to find the optimal letter-casing setting, as the repeated usage of capital letters may be more sensitive, but enlarges the vocabulary composing the training data and prohibits generalization. Our study revealed that using a lower-case setting is preferable in terms of classification performance over both upper- and lower-case character representation in the embedding layer. Therefore, as a pre-processing step we convert the data to a lower-cased representation. Characters that are not in the alphabet are mapped to a vector representation of a designated \textit{unknown} character. We denote the alphabet of characters as $V_{c}$. An embedding layer is utilized to map each one of the characters in $V_{c}$ to a lower-dimensional vector. \textcolor{black}{We set a limit of 120 characters to the input sequences. The length of the sequences varies between 80 to 150 characters, while approximately 75\% of the sequences are 120 characters or shorter. We found that a 120 character limit was sufficient for classification, as there was no clear advantage for using longer inputs. We applied padding and truncated sequences that were shorter or longer than this length limit, respectively. Finally, we evaluated several possible embedding dimensions, and chose a dimension of $d_{c} = 50$, which provided the best results. }

\begin{table}
\caption{The characters comprising the alphabet used in the proposed character-based HyperNetworks.}
\small
\setlength\tabcolsep{0pt}
\label{tab:alphabet}
\begin{tabular*}{\columnwidth}{@{\extracolsep{\fill}} ccccccccccccccc}
\hline \hline
a & b & c & d & e & f & g & h & i & j & k & l & m & n & o \\\hline
p & q & r & s & t & u & v & w & x & y & z & 0 & 1 & 2 & 3 \\\hline
4 & 5 & 7 & 8 & 9 & - & , & ; & . & ! & ? & : & ' & " & / \\\hline
\ {|} & {\_} & {@} & {\#} & {\$} & {\%} & {\textasciicircum} & {\&} & {*} & {\textasciitilde} & {`} & {+} & {-} & {=} & {\textless}\\\hline
\textgreater & ( & ) & [ & ] & \{ & \} & \textbackslash\\\hline\hline
\end{tabular*}
\end{table}

The character embedding layer is followed by two convolutional blocks. Each block is comprised of a 1-dimensional convolutional layer with 64 filters and kernel width of 7 with ReLU activation, and a 1-dimensional max pooling layer with a pool size of 4. Next, two consecutive fully-connected (FC) layers are applied with 128 and 32 perceptrons, respectively. A dropout layer is applied to the output of each FC layer, with a dropout probability of 0.5. Finally, an FC layer with a single perceptron and Sigmoid activation is applied to produce the posterior probability that the input text is hateful. We experimented with several hyper-parameters settings, including the number of convolutional layers, and selected the above set of parameter values, which provided the best performance. The complete main network, termed \textit{CharCNN},  is depicted in the lower part of Figure \ref{fig:HyperHate}.

\begin{figure*}[ht]
\centering
\includegraphics[trim={1.75cm 10cm 0cm 2cm},clip,scale=0.5]{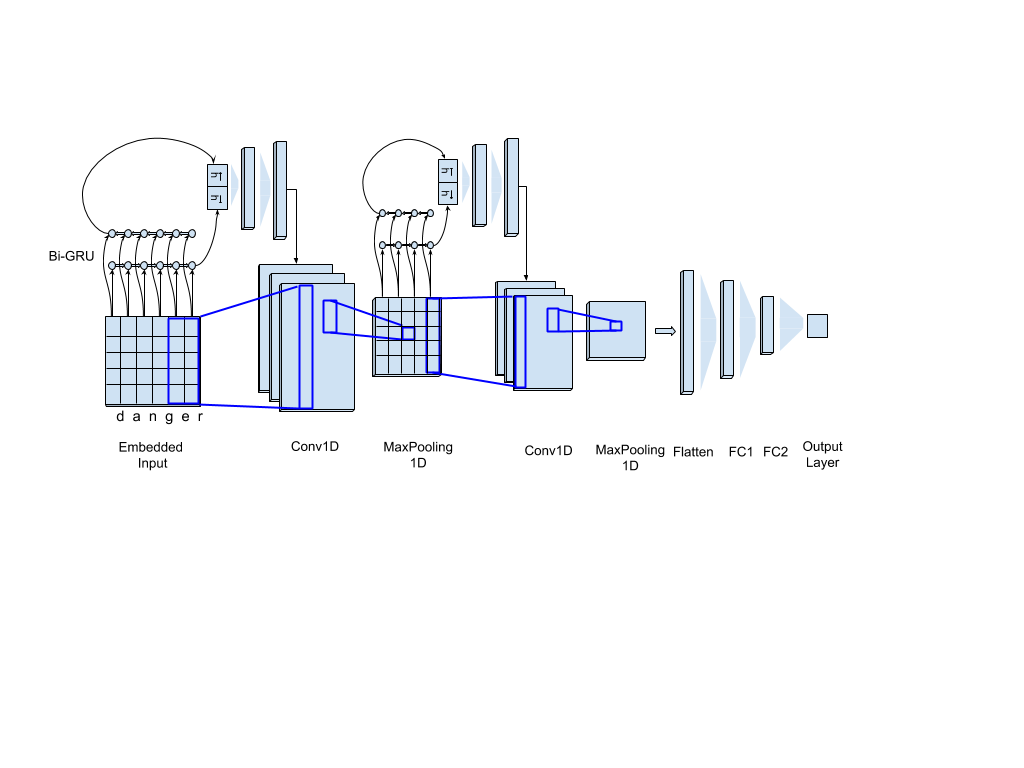}
\caption{The proposed HyperHate-Dynamic architecture:  The main network (lower part) computes the posterior probability that an input text sequence is hateful. The weights of the two 1D-convolutional layers are generated by the auxiliary network (upper part). The figure shows for illustration purposes a separate auxiliary network per convolutional layer, however, only a single auxiliary network is implemented and the weights for each convolutional layer are determined by the corresponding context vector computed by a bi-directional GRU (Bi-GRU) layer. The auxiliary network utilizes the context vector for generating input-adaptive weights for the corresponding convolutional layer, according to (\ref{eqref:static}) - (\ref{eq:context_vec}).}
\label{fig:DynamicCNN}
\end{figure*}

\paragraph{HyperHate-Static} Our first hate speech detection solution utilizes CharCNN as the main network $F(\mathbf{x};\mathbf{w})$, and a light-weight auxiliary network $G(\mathbf{z};\theta)$ for generating the weights of the  convolutional layers in the main network. The input to the auxiliary network $G$ is an embedding vector $\mathbf{z}_{j} \in \mathbb{R}^{Z}$, which is learned during network training and encodes the information of the $j$-\textit{th} convolutional layer of the main network. In our experiments, a layer embedding vector of dimension $Z=10$ was found to provide good results. Inspired by the static architecture introduced in Ha et al.~\citeyearpar{ha2016hypernetworks}, our auxiliary network $G$ consists of two non-linear layers. The first, denoted as $f_1$, performs a matrix-vector multiplication between the input embedding vector $\mathbf{z}_{j}$ and the learned matrix $\mathbf{W_{in}} \in \mathbb{R}^{(Z C_{in})\times Z}$, followed by a ReLU activation, where $C_{in}=64$ is the input depth into both convolutional layers of the main network.\footnote{For the first convolutional layer, the embedding layer output (of dimension 50) is padded by zeros to obtain a dimension of 64.} The $Z C_{in}$ dimensional vector $\mathbf{o}_1$ computed by $f_1$ is reshaped to a matrix $\mathbf{O}_1$ of dimensions  $C_{in}\times Z$ and layer $f_2$ multiplies it with a learned matrix $\mathbf{W_{out}} \in \mathbb{R}^{Z\times (kC_{out})}$, where $k=7$ is the length of each 1D convolutional kernel and $C_{out}=64$ is the number of kernels (and consequently feature maps) in both convolutional layers of the main network. The output of layer $f_2$ is a matrix $\mathbf{O}_2 \in \mathbb{R}^{C_{in}\times (kC_{out})}$ after applying element-wise ReLU activation, yielding the required $C_{in}\times k\times C_{out}$ weights  for each convolutional layer of the main network. 
The auxiliary network workflow is  described as follows: 
\begin{subequations}
\label{eqref:static}
\begin{align}
&\mathbf{o}_1=f_1(\mathbf{z}_j)=ReLU(\mathbf{W}_{in}\mathbf{z}_j) \\
&\mathbf{O}_1 = reshape(\mathbf{o}_1, [C_{in}\times Z])\\
&\mathbf{O}_2=f_2(\mathbf{O}_1)=ReLU(\mathbf{O}_1\mathbf{W}_{out}).  
\end{align}
\end{subequations}

The complete HyperHate-Static architecture is depicted in Figure \ref{fig:HyperHate}, where the auxiliary network is illustrated in the upper part of the figure: note that the figure shows for illustration purposes a separate auxiliary network per convolutional layer, however, only a single instance of the auxiliary network is implemented and the weights for each convolutional layer are determined by the unique layer embedding vector $\mathbf{z}_j$ for $j=1,2$. The total number of coefficients of the HyperHate-Static network is 76K, as detailed in Table \ref{tab:params}.  
\iffalse
In our experiments using HyperHate-Static, we use a layer embedding vector of dimension $Z=10$ and a weight generating hidden layer $f_1$ with 10  perceptrons. The rest of the model hyper-parameters are identical to CharCNN.
\fi
\iffalse The Static setting, shown in Figure~\ref{fig:HyperHate}, is similar to the static Hypernetwork described in~\citep{ha2016hypernetworks}. A network composed of two FC layers is utilized to generate parameters for CharCNN's convolutional layers. In order to produce parameters, a layer embedding vector, representing a convolutional layer within CharCNN, is fed to the generating network and projected to a \latin{$N_{dim}$} dimension, where \latin{$N_{dim}$} is the number of convolutional channels. Then, the next layer in the generating network projects the intermediate representation into the number of parameters required for the convolutional layer. We refer to this model as \textbf{HyperHate}.\fi 

\paragraph{HyperHate-Dynamic}

\iffalse In addition, we propose a dynamic setting intended to produce parameters according to layer's input. 
In this setting, each input sample {\it $x_i = [x^{1}_1, x^{1}_2,...,x^{1}_L]$} ({\it L} being the maximum sequence length) is fed into a parameter generating network, producing parameters that are better suited for that input. We refer to this setting as \textbf{HyperHate-Dynamic} (Figure~\ref{fig:DynamicCNN}). \fi

\begin{table}[t]
\caption{Size and input type of the evaluated hate speech detectors.}
\small
\centering
\label{tab:detectors}
\setlength\tabcolsep{0pt}
\begin{tabular*}{0.6\columnwidth}{@{\extracolsep{\fill}} lrl}
%\begin{tabular*}[lll]
Model & Params & Embedding Granularity\\\hline

{CharBERT\textsubscript{$BERT$}} & {130M} & {Subword and Character}\\
{RoBERTa\textsubscript{$base$}} & {125M} & {WordPiece}\\
{BERT\textsubscript{$base-uncased$}} & {110M} & {WordPiece}\\
{MobileBERT} & {24M} & {WordPiece}\\
{ALBERT\textsubscript{$large-v2$}} & {18M} & {SentencePiece}\\
{CNN-GRU} & {2.5M} & {Word}\\
%{CharCNN} & {116K} & {character}\\ 
{HyperHate-Dynamic} & {129K} & {Character}\\
{HyperHate-Static} & {76K} & {Character}\\
\hline
\end{tabular*}
\end{table}

While the HyperHate-Static model has an exceptionally low number of coefficients, it lacks the capability of adapting the main network coefficients to each different input text sequence, which can facilitate better generalization performance. Therefore, our second solution is a dynamic HyperNetwork, in which the auxiliary network has a similar architecture as in the HyperHate-Static network, excluding the replacement of the static embedding vector $\mathbf{z}_j$ by a dynamic context vector $\mathbf{h}_j$, computed online per each input text sequence and convolutional layer $j=1,2$. The context vector is computed as follows: the input sequence of vectors to the \textit{j}-th convolutional layer of the main network  {\it $X_j = [x^{j}_1, x^{j}_2,...,x^{j}_L]$} , where {\it L} is the maximal sequence length, is processed by a bidirectional GRU (Bi-GRU) layer, resulting in two final hidden states, one per direction:
\begin{subequations}
\label{eq:dynmaic}
\begin{align}
\overrightarrow{\mathbf{h}_{out}} &=  \overrightarrow{GRU}(X_{j})\\
\overleftarrow{\mathbf{h}_{out}} &= \overleftarrow{GRU}(X_{j}),
\end{align}
\end{subequations}
and the context vector is given by:
\begin{equation}
\label{eq:context_vec}
{h}_j = 
\begin{bmatrix}
\overleftarrow{\mathbf{h}_{out}}\\
\overrightarrow{\mathbf{h}_{out}}
\end{bmatrix}.
\end{equation}
The context vector $\mathbf{h}_j$ is utilized as the input to $f_1$, replacing $\mathbf{z}_j$ in Equation \ref{eqref:static}(a), for dynamically generating the main network weights. The output states of the Bi-GRU consist of 32 units per GRU direction, resulting in a context vector of dimension 64. We applied a recurrent dropout with a drop probability of 0.1, and initialized the GRU weights using a uniform glorot~\citep{glorot2010understanding} initialization.
The complete HyperHate-Dynamic architecture is depicted in Figure~\ref{fig:DynamicCNN}, where the auxiliary network is illustrated in the upper part of the figure: note that the figure shows for illustration purposes a separate auxiliary network per convolutional layer, however, only a single instance of the auxiliary network is implemented and the weights for each convolutional layer are determined by the dynamic context vector $\mathbf{h}_j$ for $j=1,2$. The total number of coefficients of the HyperHate-Dynamic network is 129K, as detailed in Table \ref{tab:params}.

\iffalse  Note that the main advantage of utilizing HyperNetworks is that the number of parameters it requires for generating weights is independent of the number of convolutional layers composing the main network, whereas a standard convolutional layer relies on a unique set of parameters for each convolutional layer.\\

\begin{table}[ht]
\caption{The number of parameters required to produce weights for a single Conv. layer in our character-based models. The Conv. layers use a $k$ width filter, and operate on $C_{in}$ and $C_{out}$ channels. The HyperHate-Static model obtain weights for Conv. layer $j$ by feeding an embedding vector $z^{j} \in \mathbb{R}^{Z}$ to a HyperNetwork comprised of two fully-connected (FC) layers. HyperHate-Dynamic utilize a bidirectional-GRU to compress the input into a context-vector, which is then received by a HyperNetwork, creating input-conditioned weights.}
\small
\centering
\label{tab:params}
\setlength\tabcolsep{0pt}
\begin{tabular*}{0.8\columnwidth}{@{\extracolsep{\fill}} lcc}
\toprule
     Model & Single Conv. Layer Params & Tot. Conv. Layers Params\\\hline\hline
{CharCNN}  & {$k \times C_{in} \times C_{out}$} & {$2 \times 28,672$}\\ 
{HyperHate-Static} & {$Z \times (Z \times C_{in}) \times (k \times C_{out})$} & {10,880}\\
{HyperHate-Dynamic} & {$W_{GRU}+2 \times Z \times (Z \times C_{in}) \times (k \times C_{out}) $}\ & {91,584}\\
\hline
\hline
\end{tabular*}
\end{table}

\fi

\begin{table*}[ht]
\centering
\caption{The number of parameters in the proposed HyperHate-Static and HyperHate-Dynamic architectures}
\resizebox{0.7\textwidth}{!}{
\small
\centering
\begin{tabular}{p{1.3cm}|p{1.5cm}|p{3cm}|p{4cm}}
%Layer & \hspace*{0.3cm}CharCNN & \hspace*{0.75cm}HyperCNN \newline \hspace*{0.75cm}-Static & \hspace*{1cm}HyperCNN \newline \hspace*{1cm}-Dynamic\\\hline\hline

\hfil Network & \hfil Layer & \hfil HyperHate-Static & \hfil HyperHate-Dynamic\\
\hline\hline
{\vfill Auxiliary } & {\vfill \hfil All} & {\hspace*{0.1cm}$Z +Z\times (Z\times C_{in})$ \newline \hspace*{0.35cm}$+Z \times (k\times C_{out})$ \newline \hspace*{1.0cm}(10,954)} & {$W_{GRU}+2\times Z\times (Z\times C_{in})$ \newline \hspace*{1cm} $+Z\times (k\times C_{out})$ \newline \hspace*{1.5cm}(64,320)}\\\hline

{\vfill \hfil Main} & {Embedding} & {\hspace*{1.0cm}$V_{c}\times d_{c}$ \newline \hspace*{1.0cm}(3,500)} & {\hspace*{1.5cm}$V_{c}\times d_{c}$ \newline \hspace*{1.5cm}(3,500)}\\\hline

{\vfill \hfil Main} & {\vfill \hfil Conv. 1} & \hfil Generated by the\newline \hspace*{0.15cm} Auxiliary Network &  \hfil Generated by the\newline  \hspace*{0.6cm} Auxiliary Network \\\hline

{\vfill \hfil Main} & {\vfill \hfil Conv. 2} & \hfil Generated by the\newline \hspace*{0.15cm} Auxiliary Network & \hfil Generated by the\newline \hspace*{0.7cm}Auxiliary Network\\\hline

{\vfill \hfil Main} & {\vfill \hfil FC 1} & {\hspace*{0.8cm}128 units \newline \hspace*{0.8cm}(57,472)} & {\hspace*{1.4cm}128 units \newline \hspace*{1.5cm}(57,472)}\\\hline
{\vfill \hfil Main} & {\vfill \hfil FC 2} & {\hspace*{1.0cm}32 units \newline \hspace*{1.0cm}(4,128)} & {\hspace*{1.4cm}32 units \newline \hspace*{1.5cm}(4,128)}\\\hline

{\vfill \hfil Main} & {\vfill \hfil FC 3} & {\hspace*{1.0cm}1 unit \newline \hspace*{1.1cm}(33)} & {\hspace*{1.5cm}1 unit \newline \hspace*{1.6cm}(33)}\\\hline
 & {\hfil Total} & {\hspace*{0.9cm}(76,087)} & {\hspace*{1.4cm}(129,453)}\\
\hline
\end{tabular}
}
\label{tab:params}
\end{table*}

\subsection{Assessing the Impact of Training Data Augmentation}
\label{sec:gen}

In this work, we assess the performance of the proposed solutions using varying amounts of training examples. The publicly available hate speech datasets are relatively small and imbalanced. As can be observed in Table~\ref{tab:datasets}, these gold-labeled (i.e., manually annotated) datasets include only thousands of examples, where a small minority of the text sequences are hate speech examples. Such a modest-sized selection can hardly represent the rich diversity of hate speech, leading to a large generalization gap  ~\citep{wullach2020towards}. Aiming at state-of-the-art text classification performance, and a high level of generalization, we opt for generating a very large number of synthetic hate- and non-hate speech text sequences for training purposes. We follow the data generation workflow proposed in~\citep{wullach2020towards}, which is based on fine-tuning the GPT-2~\citep{radford2019language} deep generative language model as described below. Compared with this previous work, we upscale the example generation effort, reaching \iffalse utilized in our study a very large-scale dataset~\citep{BANKO2001}, with\fi a total of 10 million synthetic examples, balanced over the two classes of hate and non-hate. The generated sequences are comprised of five sub-datasets of 2 million examples each, corresponding to the data distributions of the five hate speech datasets in Table~\ref{tab:datasets}. Thus, the dataset is diverse and allows to improve learning generalization.\footnote{The hate and non-hate generated sequences are available to the research community upon request from the authors.}

%\paragraph{Data Generation Workflow}
% Einat: this header was assigned a letter (a) by the style generator, but there is no (b)..

\begin{figure*}[t]
\centerline{
\includegraphics[trim={0cm 0.0cm 0cm 0cm},clip,scale=0.8]{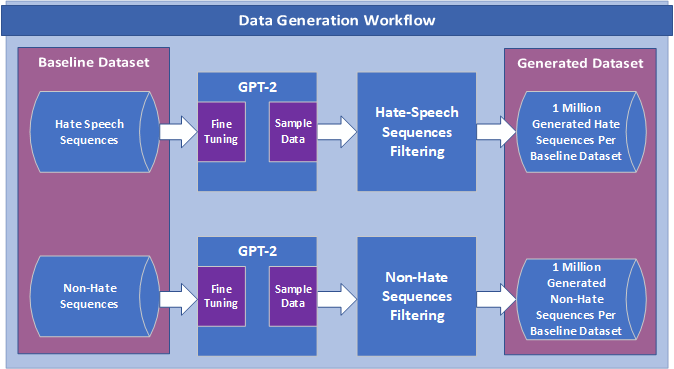}}
 \caption{The data generation workflow, adapted from~\citep{wullach2020towards}. Each dataset in Table~\ref{tab:datasets} is used to generate 2M synthetic sequences, and a total of 10M sequences using the five available datasets.}
 \label{fig:GPT_2_WF}
\end{figure*}

Concretely, the first step of the generation workflow, illustrated in Figure~\ref{fig:GPT_2_WF}, is fine-tuning two GPT-2 generative models using the training set of each dataset in Table~\ref{tab:datasets}--one using hate speech class examples and another using the non-hate class examples, resulting with a total of 10 fine-tuned GPT-2 models. We fine-tuned a separate GPT-2 model for each dataset and class (hate/non-hate) in order to learn the unique data distribution characterized by each human-labeled dataset and class. The second step of the data generation included sampling of millions of text samples from the 10 fine-tuned models. In order to ensure that the hate speech generated sequences indeed belong to each hate speech class examples, we fine-tuned BERT (base and uncased version) to classify each generated hate speech sequences as hate or non-hate. Post-filtering, a total of 10M sequences comprised our generated corpus, corresponding to 2M per dataset: 1M per each of the hate and non-hate classes. 

Importantly, we note that each one of the datasets in Table~\ref{tab:datasets} was randomly split into disjoint training (80\%) and test (20\%) sets. \iffalse as follows: each dataset was randomly split to a training set (80\%) and a held-out testing set (20\%).\fi The data generation processes and all the training procedures described in Section~\ref{sec:results} were performed using the training sets, whereas hate detection performance was evaluated exclusively using the corresponding held-out testing sets.

\section{Performance Evaluation}
\label{sec:results}

This Section provides detailed performance evaluation of the proposed character-based HyperNetwork solutions compared to state-of-the-art deep learning architectures of hate detection, including the word-based CNN-GRU hate speech detector by Zhang et al.~\citeyearpar{zhang2018detecting}, and several pretrained Transformer-based models, namely BERT, RoBERTa and ALBERT. Concretely, we experiment with the large variant of the ALBERT model, which was shown to yield comparable results to BERT-base, but is much smaller, including only 18M parameters versus 110M. Both BERT and ALBERT were pretrained using the same data--the BookCorpus and the whole of English Wikipedia, while RoBERTa has been trained using additional data. In addition, we experiment with the newly introduced model of CharBERT, which incorporates character representations into BERT. We train our proposed character-based solutions and the CNN-GRU, and similarly fine-tune the pretrained models, using the available labeled training examples, where we test and compare the performance of the different methods on the held-out labeled examples. We further evaluate learning performance using increasing amounts of synthetic examples, augmenting the labeled data with up to 2M additional generated examples per dataset. While all the models may benefit from this procedure~\citep{wullach2020towards}, we expect the extended training to improve significantly the character-based models by exposing them to relevant language statistics.

Placing emphasis on learning generalization, our analysis includes two types of experiments, which we refer to as \textit{intra-domain} and \textit{cross-domain} hate speech detection. In the {\it intra-domain} experiments we follow the common practice of applying the DL detectors to each dataset independently: the performance of each DL detector is measured using the held-out gold-labeled test examples of every dataset $D_i$, having trained the detector with either the baseline training set, or with the augmented training set available for $D_i$, across the range of 0-2M generated sequences per dataset. In addition, we formed a \textit{combined} dataset comprised of the union of all datasets (having joined all training and test sets, respectively). The {\it cross-domain} experiments aim to evaluate the realistic condition of data distribution shift, where the hate detector is applied to text sequences sampled from a data distribution that is different from the one used for training~\citep{wiegand2019detection,wullach2020towards,wullachEMNLP21}. In these experiments, having the detector trained using the training set of dataset $\mathcal{D}_s$, it is evaluated on the held-out test set of dataset $\mathcal{D}_t\neq \mathcal{D}_s$. Again, we trained each DL detector using either the baseline or the augmented training set available for dataset $D_s$, across the range of 0-2M generated sequences per dataset. 

We report classification performance with respect to the hate class in term of Recall, Precision and \textit{F}1 scores. Recall corresponds to the proportion of true hate speech examples that were automatically identified as hate speech, and Precision is the proportion of correct predictions within the examples identified as hate speech by the detector. \textit{F}1 is the harmonic mean of those measures, assigning them equal importance. Let us note that assuming that only a small proportion of the data is automatically identified as hate speech, false positive predictions may be tracked relatively easily by means of further human inspection, whereas it is impossible to track false negatives at scale. Therefore, increasing recall by means of improved generalization is of great importance in practice.  \\

\subsection{Implementation Details}

We implemented all the DL models in TensorFlow~\citep{tensorflow2015-whitepaper} and utilized the NVIDIA K-80 GPU for training and testing. In training the proposed HyperHate models, we applied the Adam optimizer with a mini-batch size of 32 samples and an early stopping mechanism, minimizing the binary cross-entropy loss. Additional  implementation choices of the proposed HyperNetwork models are discussed in Section~\ref{sec:approach}.\footnote{We make our code available at https://github.com/tomerwul/CharLevelHyperNetworks.}

The CNN-GRU detector was trained using the Adam optimizer~\citep{Goodfellow-et-al-2016}, minimizing the binary cross-entropy loss, with early stopping and mini-batch size of 32. The 1D-Convolutional layer includes 100 filters with kernel size of 4, the 1D-MaxPooling layer with pool size of 4 and the GRU output dimension of 100.

Our implementation of BERT and the other pretrained models makes use of the popular HuggingFace Transformers repository.\footnote{https://github.com/huggingface/transformers} \iffalse which provides a variety of pre-trained transformer-based models.\fi In all cases, we fine-tuned the pretrained model in its uncased setting, with an additional feed-forward layer on top of the final output of the [CLS] token, which yields a probability distribution for a binary classification task. Following a validation step, we fine-tuned the models for 3 epochs with a mini-batch size of 32, and used Adam optimizer with an initial learning rate of 2e-5, performing 1000 warm up steps prior to reaching the initial learning rate. We set a fixed sentence length of 30 tokens, padding input sentences that are shorter and truncating longer sentences. An attention mask was employed to avoid including the padded tokens in the self-attention calculations.
\iffalse Gururangan et al.~\citep{gururACL2020} suggests sampling examples that are similar to the classification training data from a task-related unlabeled corpus, and adapt the model using those examples. Here, we use generated labeled examples, which are artificial and noisy to a certain extent, for extending the gold-labeled, yet biased and small, existing datasets of hate speech. We evaluate the impact of train set augmentation in fine-tuning BERT with increasing amounts of our generated data. \fi
Our implementation specifications of ALBERT are similar to those utilized for fine-tuning BERT, with the exception of using a text tokenizer that is suited for the ALBERT model.\\

\begin{figure*}[p]
\centerline{
\includegraphics[width=24cm]{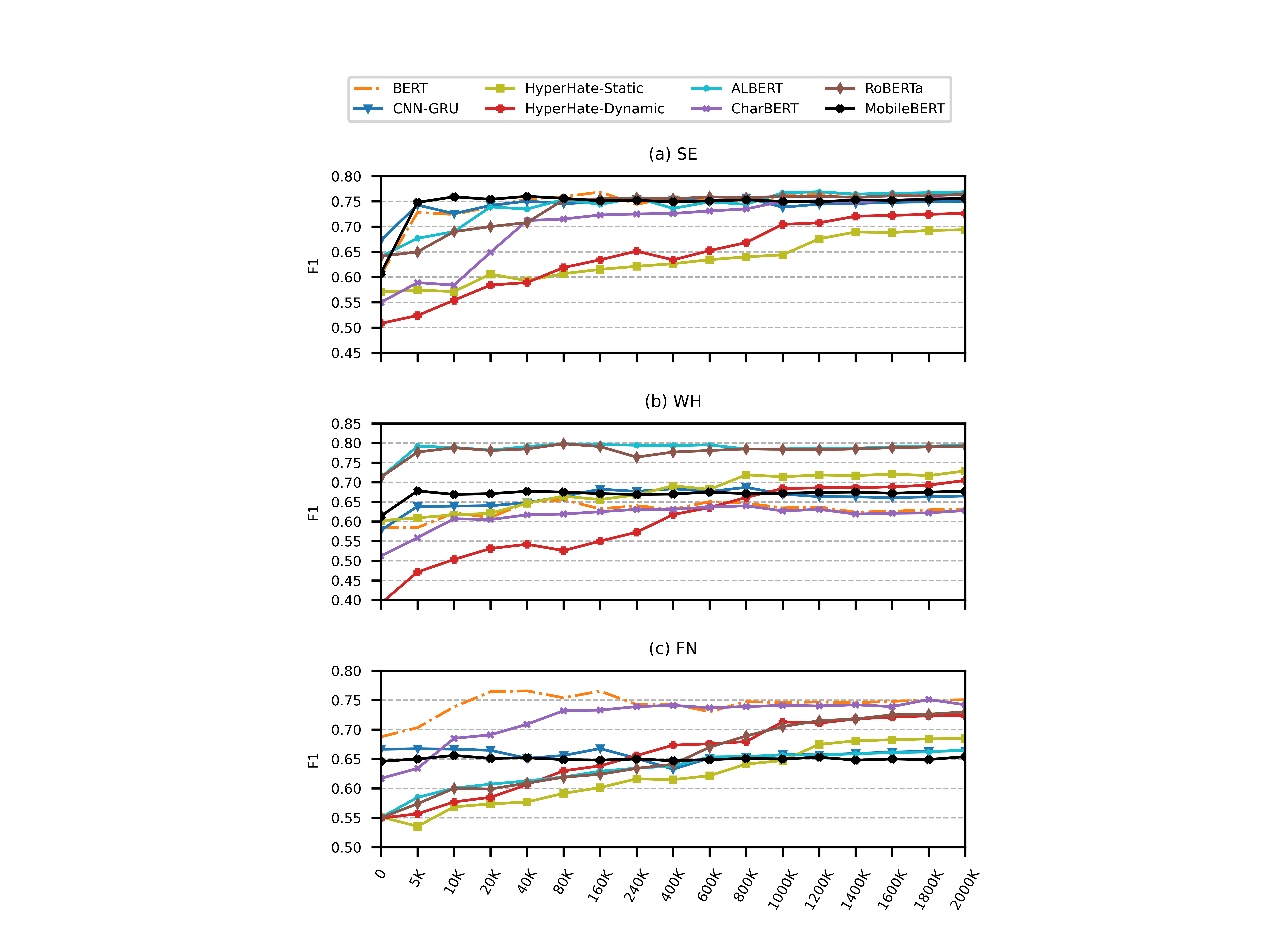}}
 \caption{Intra-dataset results (part 1/2): hate-class F1, using increasing amounts of 0-2M generated sequences added as data augmentation.}
  \label{fig:intra-Curves}
\end{figure*}

\begin{figure*}[p]
\centerline{
\includegraphics[trim={0cm 0cm 0cm 0cm},clip,scale=1.4]{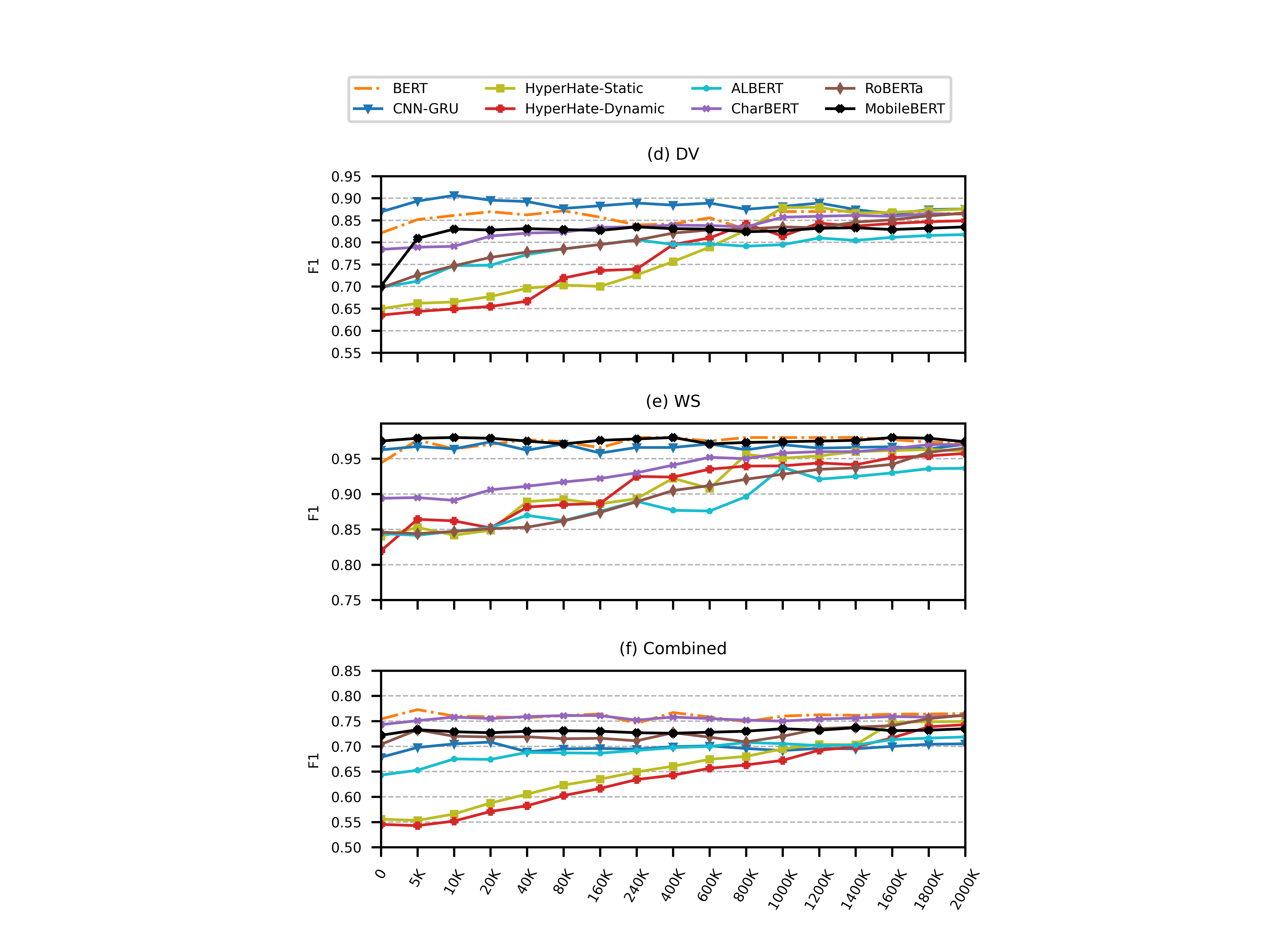}}
 \caption{Intra-dataset results (part 2/2): hate-class F1, using increasing amounts of 0-2M generated sequences added as data augmentation.}
  \label{fig:intra-Curves_b}
\end{figure*}

\iffalse
\begin{figure*}[t]
\centerline{
\includegraphics[trim={1cm 0.5cm 1.25cm 0.15cm},clip,scale=1.2]{cross_curves_4on1.png}}
 \caption{Combined cross-domain results. Plots (a)-(d) show the results of training the detectors on a combined dataset consists of four hate speech datasets (augmented with the corresponding generated sequences) and evaluated on a test set taken from a fifth dataset. }
  \label{fig:4on1-Curves}
\end{figure*}
\fi

\begin{figure*}[p]
\centerline{
\includegraphics[trim={1cm 0.0cm 1.25cm 0cm},clip,scale=1.4]{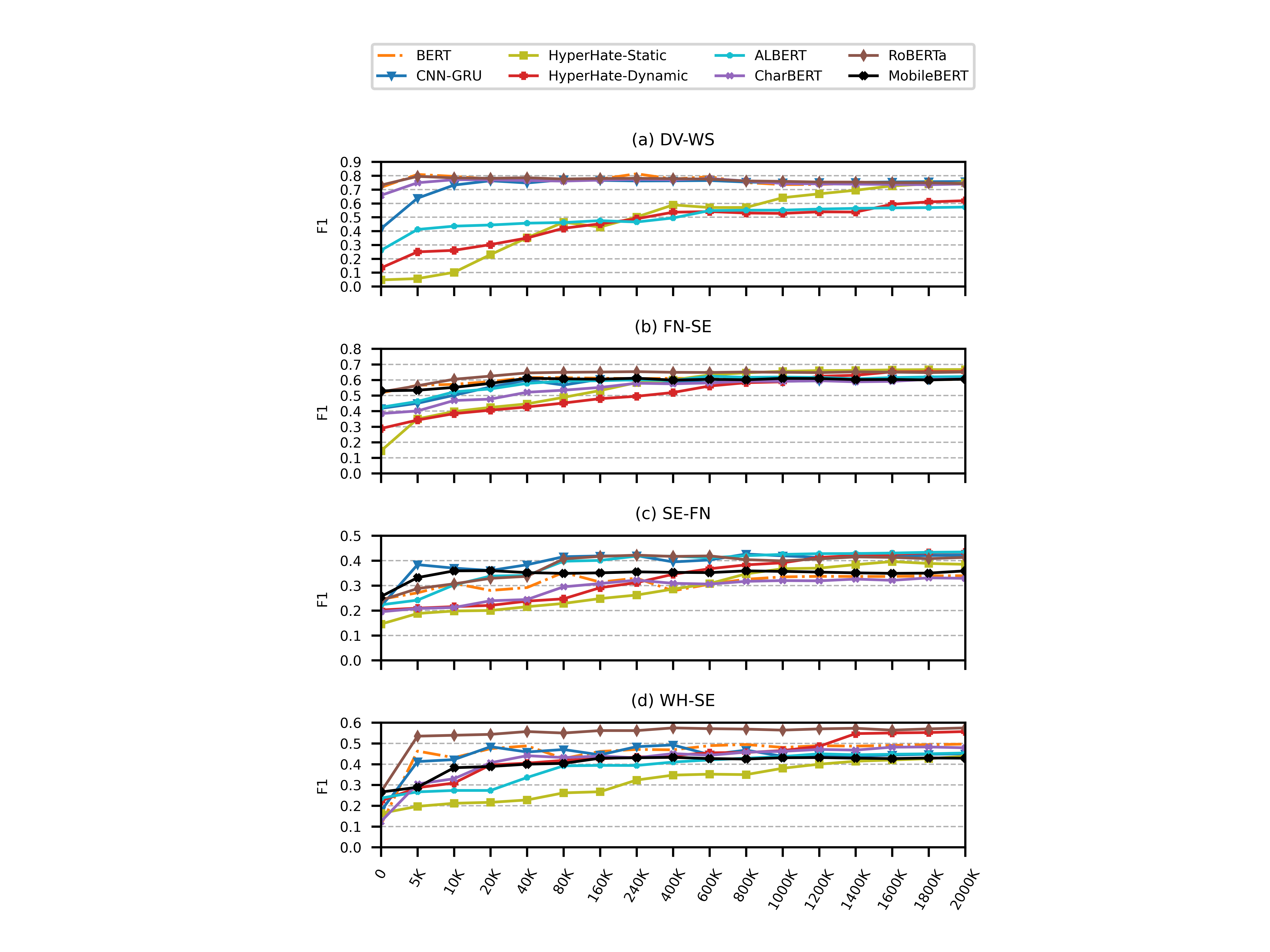}}
 \caption{Cross-dataset results (part 1/2): hate-class F1, using increasing amounts of 0-2M generated sequences added as data augmentation.
 The detectors were trained on the left dataset, and tested on the held-out examples of the right dataset, of each dataset pair. The best {\it F}1 result per dataset pair are highlighted in boldface.}
  \label{fig:cross-Curves}
\end{figure*}

\begin{figure*}[p]
\centerline{
\includegraphics[trim={1cm 0.0cm 1.25cm 0cm},clip,scale=1.4]{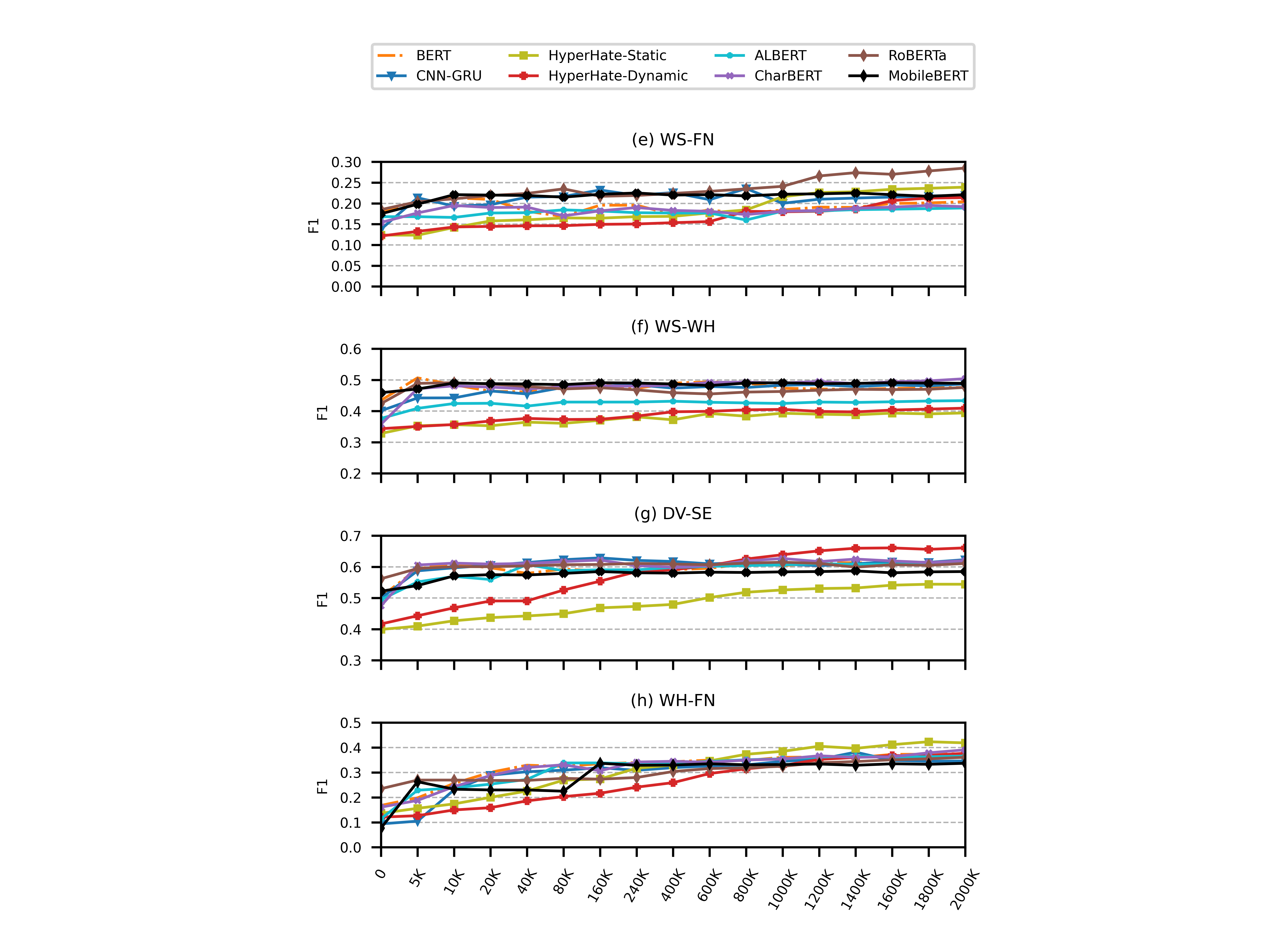}}
 \caption{Cross-dataset results (part 2/2): hate-class F1, using increasing amounts of 0-2M generated sequences added as data augmentation.
 The detectors were trained on the left dataset, and tested on the held-out examples of the right dataset, of each dataset pair. The best {\it F}1 result per dataset pair are highlighted in boldface.}
  \label{fig:cross-Curves_b}
\end{figure*}

\subsection{Intra-domain Experiments: Results} 

Figures~\ref{fig:intra-Curves} and~\ref{fig:intra-Curves_b} report the results of the intra-domain experiments in terms of $F1$ scores, computed with respect to the {\it hate} class. The two figures present the results for the various methods, trained and tested using each of the five source datasets and their union, denoted as the \textit{combined} dataset. The classifiers' performance using no (0) augmentation corresponds to training using merely the gold-labeled training portion (80\%) of the source datasets. The figures show the results of augmenting this gold-labeled data with increasing amounts of up to 2M class-balanced generated examples.

\textcolor{black}{Several trends are observed in the results. First, we observe that the larger token-based models, including BERT, RoBERTa, and ALBERT, as well as CNN-GRU, dominate the proposed character-based HyperNetwork models when the amount of labeled examples is small (i.e., with no data augmentation). Indeed, it has been previously shown that the models that are pretrained using large amounts of text are advantageous given limited labeled examples~\citep{wullachEMNLP21}. It is further observed that, for most of the models and datasets, a boost in the performance is achieved once as few as 5-10K generated examples are added to the gold-labeled data. As the original datasets contain only a few thousands of gold-labeled examples, adding a similar amount of synthetic data increases data diversity, roughly doubling the train set size, where this has a large impact on learning performance.} 

\textcolor{black}{We further observe in Figures~\ref{fig:intra-Curves}  and~\ref{fig:intra-Curves_b} that the respective $F1$ Performance levels of the token-based models typically continues to rise slowly or stabilizes as more generated labeled examples are provided. In comparison, the HyperNetwork models that are inferior initially leverage the additional generated training data, where the peak performance using these models is obtained in the range of 1-2 million generated examples. That is, large amounts of data is required to train these models, which were not pretrained, to identify hate- and non-hate related character sequences. Crucially, the figures show that the gap in performance between the large token-based and compact character-based models narrows down with increased levels of data augmentation.}

Detailed results including precision, recall and \textit{F}1 scores using full vs. no augmentation per method and dataset are given in Table~\ref{tab:intra_dataset_results} (Appendix~\ref{sec:appendix_results}). \textcolor{black}{The best results in terms of \textit{F}1 are highlighted in boldface per dataset in the table.}
As detailed in Table~\ref{tab:intra_dataset_results}, all of the methods benefit from data augmentation by 2M generated sequences. Mainly, we observe significant and consistent gains in recall with data augmentation for all methods. The recall improvements are more pronounces for the HyperNetwork models, peaking at +153.58$\%$ for the HyperHate-Dynamic model and StormFront (WH) dataset. Notably, the introduction of large amounts of synthetic text sequences may introduce a drift in the data distribution, or noise. As shown, in some cases the precision scores are negative, with a minimum of -31.54$\%$ for the ALBERT model and Founta (FN) dataset.
Nevertheless, precision declines to a lesser degree compared to the high gains in recall. And, in some cases, both precision and recall benefit from data augmentation (e.g., see the results for most methods on the WS dataset). Overall, hate detection improves with data augmentation, as reflected by the \textit{F}1 scores, which are overall positive. We note  that while \textit{F}1 performances rise between $\sim$0-20\% for the larger models and the individual datasets, the respective improvements for the character-based HyperNetworks are significantly higher, ranging between $\sim$14-70\%. 

\textcolor{black}{Overall, as shown in Table~\ref{tab:intra_dataset_results}, following data augmentation, the HyperNetwork architectures achieve preferable or competitive results compared with some of the large networks following data augmentation, e.g., achieving best results along with CNN-GRU on the Davidson (DV) test set; outperforming CNN-GRU, BERT, CharBERT and MobileBERT on StormFront (WH), CNN-GRU and ALBERT on the Founta (FN) dataset; and, the model of ALBERT on the Waseem (WS) dataset. Also, these models achieve comparable performance to the large pretrained token-based models, while slightly outperforming ALBERT, MobileBERT and CNN-GRU, on the combined dataset. This result has striking implications considering the difference in size of these types of models.}

\subsection{Cross-domain Experiments: Results} 

The results of the cross-domain experiments are displayed in Figures~\ref{fig:cross-Curves} and~\ref{fig:cross-Curves_b} for eight representative source-target dataset pairs in terms of \textit{F}1 scores. In the cross-domain setup, the models are trained using the gold-labeled train set of a source dataset $\mathcal{D}_s$, being augmented with synthetic examples generated from the same train data distribution of $\mathcal{D}_s$, and are then evaluated over the held-out test set of a target dataset $\mathcal{D}_t$. The figures show the results of augmenting the gold-labeled training data with increasing amounts of 0-2M class-balanced generated examples. 

The cross-domain setup is more challenging learning-wise and is more realistic than the intra-domain setup as it introduces a shift in data distribution. Accordingly, the overall results are dramatically lower compared with the intra-dataset experiments. The extent of drop in performance varies across the dataset pairs, depending on the gap in their characteristics.

Overall, we observe consistent significant performance gains with data augmentation for all methods. In particular, there is a steep rise of recall in most of the experiments. In terms of precision scores, there are substantial gains in some cases due to data augmentation (peaking at +332.25$\%$ using HyperNetwork-static, and +229.9$\%$ using ALBERT), where in a few cases precision decreased following augmentation (reaching a minimum of -48.21$\%$ in one of the experiments, namely the ALBERT model and WS-FN pair; yet, also in that case the overall {\it F}1 performance improved due to the larger boost in recall).  The boost in recall and overall {\it F}1 performance is especially prominent for the HyperNetwork models, e.g., reaching up to +2925$\%$ improvement in recall and +1477$\%$ in {\it F}1 for the HyperHate-Static model and DV-WS pair. 

Detailed results, including precision, recall and \textit{F}1 scores are given in Table~\ref{tab:cross_dataset_results} (Appendix~\ref{sec:appendix_results}), comparing the performance per dataset and method with and without data augmentation by 2M sequences. The best hate detection performance across methods is highlighted in boldface in the table for each dataset pair. \textcolor{black}{It is an overwhelming result that the compact HyperNetwork models reach better performance compared to the larger models in some cases following data augmentation. Specifically, the dynamic HyperNetwork variant yields the best performance score for the DV-SE and the SE-FN pairs (on par with ALBERT, for the latter pair). The Static HyperNetwork model provides the best performance for the FN-SE and WH-FN pairs and is second-best to RoBERTa on the WH-SE pair. In general, it is observed in the table that while all models improve from data augmentation, both of the HyperNetwork configurations gain more in performance following data augmentation in almost all of the experiments as compared to the larger token-based models, both in terms of recall and precision. } 

\textcolor{black}{Overall, the static and dynamic HyperNetwork variants perform comparably. The best choice of model may depend on the characteristics of the data, and can be determined using validation experiments, following the common practice in machine learning.} 

To conclude, the reported results demonstrate strong performance of the proposed character-level HyperNetwork solutions on hate detection across data distributions. Presumably, these compact models are less expressive compared to large pretrained language models such as BERT, RoBERTa and CharBERT. \textcolor{black}{Yet, they achieved superior or competitive performance in a substantial portion of our cross-dataset experiments. We conjecture that character-level processing may be less overfitting to the language statistics observed in training, being more adaptive to word variants and inflections which vary across datasets.}

\iffalse In terms of recall scores, we observe consistently significant gains with data augmentation for all methods, peaking at +2925.0$\%$ for the HyperHate-Static model and DV-WS pair. In terms of precision scores, the gains due to data augmentation are typically lower, peaking at +332.25$\%$ and in few cases even negative with a minimum of -48.21$\%$ for the ALBERT model and WS-FN pair.\fi

\section{Conclusions}

In this paper, we proposed light-weight network architectures that make use of efficient mechanisms of weight sharing, namely HyperNetworks, as the backbone of character-based models trained to detect hate speech. Concretely, we described and experimented with a static auxiliary network where the generated weights are fixed during inference, and proposed a novel dynamic auxiliary network that produces input-conditioned weights. 

We evaluated hate detection performance  using gold-labeled examples of five public datasets, both using intra- and cross-dataset settings, while performing data augmentation using varying amounts of generated synthetic data. \textcolor{black}{Alongside the suggested methods, we reported hate detection results using BERT and RoBERTa, popular pretrained transfomer-based models which yield high performance of text classification but are large and computationally expensive. We further evaluated CharBERT, a variant of BERT that incorporates character embeddings, and CNN-GRU, a well-performing token-based hate detection model.} The results indicate that the proposed methods, while producing inferior results at first when no augmentation is added, can effectively leverage additional synthetic data and achieve competitive results compared to these state-of-the-art models. Furthermore, we obtained superior results using the proposed architectures in \textcolor{black}{some cases in} the challenging cross-dataset setup, simulating the data shift that typically occurs in practice across domains and over time. Thus, the proposed character-level HyperNetwork solutions present a combination of high performance as well as compactness, being smaller by orders of magnitude compared with state-of-the-art architectures, making them suitable for hate detection in general, and applicable for low-memory end devices in particular.

\textcolor{black}{The proposed character-based architectures are  language independent. We believe that these methods will prove particularly advantageous in morphology-rich languages, where character sequences can model the variation that is observed in social media text, as well as related word forms. This would require relevant resources of labeled data, and mechanisms of generating additional high-quality relevant texts in the target language~\citep{deVriesACL21,fengACL21}. Another interesting direction to explore is the use of data that is automatically labeled using semi-supervised classification~\citep{sarkarEMNLP21}, as alternative or in addition to synthetically generated data, in training the character-based hate detection models.}



\section*{Acknowledgement}
This research was supported partly by Facebook Content Policy Research on Social Media
Platforms Research Award.

%\begin{scriptsize}
\bibliographystyle{model5-names}
\bibliography{main}
%\end{scriptsize}
%% else use the following coding to input the bibitems directly in th

%\begin{thebibliography}{00}


%\bibitem{}

\section*{Appendix}
\appendix
\section{Detailed Intra-dataset and Cross-dataset results}
\label{sec:appendix_results}

\begin{table*}[h]
\centering
\caption{Intra-domain experimental results, demonstrating the impact of augmenting the training data by adding 2M generated sequences per dataset.}
\label{tab:intra_dataset_results}
\resizebox{\textwidth}{!}{
\begin{tabular}{ll|ccr|ccr|ccr}
     & & \multicolumn{3}{c}{DV} & \multicolumn{3}{c}{WH} & \multicolumn{3}{c}{SE}\\
     Architecture & & Baseline & Augmented & (\%) & Baseline & Augmented & (\%) & Baseline & Augmented & (\%)\\ \hline
    \multirow{1}{*}{CNN-GRU} & Prec. & {0.922} & {0.861} & {-6.61} & {0.748} & {0.724} & {-3.20} & {0.855} & {0.709} & {-17.07}\\
    & Recall & {0.822} & {0.891} & {+8.39} & {0.471} & {0.615} & {+30.50} & {0.555} & {0.797} & {+43.60}\\
    & F1 & {0.869} & {\textbf{0.875}} & {+0.76} & {0.578} & {0.665} & {+15.05} & {0.673} & {0.750} & {+11.49}\\\hline
    
    \multirow{1}{*}{BERT} & Prec. & {0.981} & {0.859} & {-12.43} & {0.609} & {0.709} & {+16.42} & {0.696} & {0.716} & {+2.87}\\
    & Recall & {0.706} & {0.874} & {+23.79} & {0.562} & {0.570} & {+1.42} & {0.535} & {0.826} & {+54.39}\\
    & F1 & {0.821} & {0.866} & {+5.52} & {0.584} & {0.631} & {+8.10} & {0.604} & {0.767} & {+9.13}\\\hline
    
    \multirow{1}{*}{CharBERT} & Prec. & {0.979} & {0.863} & {-11.84} & {0.614} & {0.704} & {+14.65} & {0.691} & {0.727} & {
    5.20}\\
    & Recall & {0.694} & {0.866} & {+24.78} & {0.558} & {0.572} & {+2.50} & {0.541} & {0.831} & {+53.60}\\
    & F1 & {0.812} & {0.864} & {+6.43} & {0.584} & {0.631} & {+7.95} & {0.606} & {\textbf{0.775}} & {+27.79}\\\hline
    
    \multirow{1}{*}{RoBERTa} & Prec. & {0.824} & {0.864} & {+4.85} & {0.809} & {0.775} & {-4.20} & {0.64} & {0.715} & {+11.71}\\
    & Recall & {0.605} & {0.870} & {+43.80} & {0.637} & {0.810} & {+27.15} & {0.642} & {0.821} & {+27.88}\\
    & F1 & {0.697} & {0.866} & {+24.26} & {0.712} & {0.792} & {+11.13} & {0.641} & {0.764} & {+19.24}\\\hline
    
    \multirow{1}{*}{ALBERT} & Prec. & {0.824} & {0.852} & {+3.39} & {0.809} & {0.776} & {-4.07} & {0.640} & {0.717} & {+12.03}\\
    & Recall & {0.605} & {0.786} & {+29.91} & {0.637} & {0.813} & {+27.62} & {0.642} & {0.829} & {+29.12}\\
    & F1 & {0.697} & {0.817} & {+17.19} & {0.712} & {\textbf{0.794}} & {+11.40} & {0.641} & {0.768} & {+19.96}\\\hline
    
    \multirow{1}{*}{MobileBERT} & Prec. & {0.815} & {0.828} & {+1.59} & {0.765} & {0.712} & {-6.92} & {0.680} & {0.699} & {+2.79}\\
    & Recall & {0.818} & {0.843} & {+37.07} & {0.514} & {0.646} & {+25.68} & {0.550} & {0.825} & {+50.00}\\
    & F1 & {0.701} & {0.835} & {+19.17} & {0.614} & {0.677} & {+10.16} & {0.608} & {0.756} & {+24.44}\\\hline
    
    \multirow{1}{*}{HyperHate-Static} & Prec. & {0.904} & {0.861} & {-4.75} & {0.898} & {0.751} & {-16.36} & {0.654} & {0.596} & {-8.86}\\
    & Recall & {0.507} & {0.891} & {+75.73} & {0.453} & {0.708} & {+56.29} & {0.506} & {0.830} & {+64.03}\\
    & F1 & {0.649} & {\textbf{0.875}} & {+34.80} & {0.602} & {0.728} & {+21.03} & {0.570} & {0.693} & {+21.60}\\\hline
    
    \multirow{1}{*}{HyperHate-Dynamic} & Prec. & {0.850} & {0.832} & {-2.11} & {0.750} & {0.741} & {-1.20} & {0.682} & {0.613} & {-10.11}\\
    & Recall & {0.507} & {0.867} & {+71.00} & {0.265} & {0.672} & {+153.58} & {0.405} & {0.891} & {+120.00}\\
    & F1 & {0.635} & {0.849} & {+33.69} & {0.391} & {0.704} & {+79.97} & {0.508} & {0.726} & {+42.91}\\\hline\hline
\end{tabular}}

\resizebox{\textwidth}{!}{
\begin{tabular}{ll|ccr|ccr|ccr}
     & & \multicolumn{3}{c}{FN}  & \multicolumn{3}{c}{WS} & \multicolumn{3}{c}{Combined}\\
    Architecture & & Baseline & Augmented & (\%) & Baseline & Augmented & (\%) & Baseline & Augmented & (\%)\\ \hline
    \multirow{1}{*}{CNN-GRU} & Prec. & {0.821} & {0.613} & {-25.33} & {0.994} & {0.979} & {-1.50} & {0.827} & {0.666} & {-10.31}\\
    & Recall & {0.561} & {0.724} & {+29.05} & {0.933} & {0.962} & {+3.10} & {0.575} & {0.749} & {+88.99}\\
    & F1 & {0.666} & {0.663} & {-0.39} & {0.962} & {0.970} & {+0.82} & {0.678} & {0.705} & {+34.83}\\\hline
    
    \multirow{1}{*}{BERT} & Prec. & {0.730} & {0.657} & {-10.00} & {0.944} & {0.968} & {+2.54} & {0.816} & {0.694} & {-14.95}\\
    & Recall & {0.650} & {0.875} & {+34.61} & {0.944} & {0.977} & {+3.49} & {0.701} & {0.852} & {+21.54}\\
    & F1 & {0.687} & {0.750} & {+26.79} & {0.944} & {0.972} & {+3.01} & {0.754} & {\textbf{0.764}} & {+3.93}\\\hline
    
    \multirow{1}{*}{CharBERT} & Prec. & {0.721} & {0.663} & {-8.04} & {0.923} & {0.978} & {+5.95} & {0.794} & {0.685} & {-13.72}\\
    & Recall & {0.623} & {0.870} & {+39.64} & {0.942} & {0.950} & {+0.84} & {0.687} & {0.844} & {+22.85}\\
    & F1 & {0.668} & {\textbf{0.752}} & {+12.58} & {0.932} & {0.963} & {+3.36} & {0.736} & {0.756} & {+2.66}\\\hline
    
    \multirow{1}{*}{RoBERTa} & Prec. & {0.897} & {0.870} & {-3.01} & {0.841} & {0.997} & {+18.54} & {0.781} & {0.731} & {-6.40}\\
    & Recall & {0.397} & {0.629} & {+58.43} & {0.847} & {0.936} & {+10.50} & {0.642} & {0.797} & {+24.14}\\
    & F1 & {0.550} & {0.730} & {+32.65} & {0.843} & {0.965} & {+14.40} & {0.704} & {0.762} & {+8.21}\\\hline
    
    \multirow{1}{*}{ALBERT} & Prec. & {0.897} & {0.614} & {-31.54} & {0.841} & {0.943} & {+12.12} & {0.845} & {0.673} & {-20.35}\\
    & Recall & {0.397} & {0.726} & {+82,87} & {0.847} & {0.930} & {+9.79} & {0.519} & {0.771} & {+48.55}\\
    & F1 & {0.550} & {0.665} & {+20.87} & {0.843} & {0.936} & {+10.95} & {0.643} & {0.718} & {+11.76}\\\hline
    
    \multirow{1}{*}{MobileBERT} & Prec. & {0.814} & {0.752} & {-7.61} & {0.979} & {0.99} & {+1.12} & {0.786} & {0.761} & {-3.18}\\
    & Recall & {0.536} & {0.579} & {+8.02} & {0.972} & {0.959} & {-1.33} & {0.669} & {0.712} & {6.42}\\
    & F1 & {0.646} & {0.654} & {+1.22} & {\textbf{0.975}} & {0.974} & {-0.12} & {0.722} & {0.735} & {1.78}\\\hline
    
    \multirow{1}{*}{HyperHate-Static} & Prec. & {0.742} & {0.624} & {-15.90} & {0.882} & {0.961} & {+8.95} & {0.766} & {0.687} & {-10.31}\\
    & Recall & {0.439} & {0.759} & {+72.89} & {0.802} & {0.961} & {+19.82} & {0.436} & {0.824} & {+88.99}\\
    & F1 & {0.551} & {0.684} & {+24.16} & {0.840} & {0.961} & {+14.39} & {0.555} & {0.749} & {+34.83}\\\hline
    
    \multirow{1}{*}{HyperHate-Dynamic} & Prec. & {0.821} & {0.644} & {-21.55} & {0.796} & {0.954} & {+19.84} & {0.761} & {0.684} & {-10.11}\\
    & Recall & {0.413} & {0.827} & {+100.24} & {0.845} & {0.961} & {+13.72} & {0.425} & {0.813} & {+91.29}\\
    & F1 & {0.549} & {0.724} & {+31.76} & {0.819} & {0.957} & {+16.79} & {0.545} & {0.742} & {+36.21}\\\hline
\end{tabular}}
\end{table*}

\begin{table*}[t]
\centering
\caption{Cross-domain experimental results, demonstrating the impact of augmenting the training data by adding 2M generated sequences per dataset.}
\label{tab:cross_dataset_results}
\resizebox{\textwidth}{!}{
\begin{tabular}{ll|ccr|ccr|ccr|ccr}
     & & \multicolumn{3}{c}{WH-FN} & \multicolumn{3}{c}{DV-SE} & \multicolumn{3}{c}{WH-SE} & \multicolumn{3}{c}{FN-SE}\\
    Architecture & & Baseline & Augmented & (\%) & Baseline & Augmented & (\%) & Baseline & Augmented & (\%) & Baseline & Augmented &(\%)\\ \hline
    \multirow{1}{*}{CNN-GRU} & Prec. & {0.410} & {0.408} & {-0.5} & {0.582} & {0.500} & {-14.1} & {0.614} & {0.600} & {-2.3} & {0.667} & {0.572} & {-14.2}\\
    & Recall & {0.053} & {0.301} & {+467.9} & {0.438} & {0.827} & {+88.8} & {0.102} & {0.360} & {+252.9} & {0.304} & {0.640} & {+110.5}\\
    & F1 & {0.093} & {0.346} & {+269.1} & {0.499} & {0.623} & {
    +24.7} & {0.174} & {0.450} & {+157.2} & {0.417} & {0.604} & {+44.6}\\\hline
    
    \multirow{1}{*}{BERT} & Prec. & {0.525} & {0.730} & {+39.0} & {0.563} & {0.534} & {-5.2} & {0.639} & {0.630} & {-1.4} & {0.641} & {0.572} & {-10.8}\\
    & Recall & {0.100} & {0.253} & {+153.0} & {0.456} & {0.716} & {+57.0} & {0.060} & {0.410} & {+583.3} & {0.433} & {0.639} & {+47.6}\\
    & F1 & {0.168} & {0.375} & {+123.7} & {0.503} & {0.611} & {+21.4} & {0.109} & {0.496} & {+352.8} & {0.516} & {0.603} & {+16.8}\\\hline
    
    \multirow{1}{*}{CharBERT} & Prec. & {0.521} & {0.719} & {+38.00} & {0.559} & {0.537} & {-3.93} & {0.642} & {0.640} & {-0.31} & {0.638} & {0.575} & {-9.87}\\
    & Recall & {0.090} & {0.248} & {+175.55} & {0.458} & {0.719} & {+56.98} & {0.110} & {0.411} & {+273.63} & {0.429} & {0.641} & {+49.41}\\
    & F1 & {0.153} & {0.368} & {+140.27} & {0.503} & {0.614} & {+22.11} & {0.187} & {0.500} & {+166.50} & {0.513} & {0.606} & {+18.16}\\\hline
    
    \multirow{1}{*}{RoBERTa} & Prec. & {0.198} & {0.361} & {+53.67} & {0.664} & {0.527} & {-20.63} & {0.582} & {0.586} & {+0.68} & {0.651} & {0.563} & {-13.51}\\
    & Recall & {0.289} & {0.241} & {+263.63} & {0.487} & {0.727} & {+49.28} & {0.173} & {0.564} & {+226.01} & {0.437} & {0.771} & {+76.43}\\
    & F1 & {0.235} & {0.361} & {+53.67} & {0.561} & {0.611} & {+8.74} & {0.266} & {\textbf{0.574}} & {+115.50} & {0.522} & {0.650} & {+24.44}\\\hline
    
    \multirow{1}{*}{Albert} & Prec. & {0.127} & {0.419} & {+229.9} & {0.550} & {0.535} & {-2.7} & {0.440} & {0.603} & {+91.7} & {0.669} & {0.666} & {-13.0}\\
    & Recall & {0.104} & {0.325} & {+212.5} & {0.450} & {0.712} & {+58.2} & {0.162} & {0.364} & {+37.0} & {0.311} & {0.676} & {+114.1}\\
    & F1 & {0.114} & {0.366} & {+220.1} & {0.495} & {0.610} & {+23.4} & {0.236} & {0.453} & {+124.7} & {0.424} & {0.621} & {+46.3}\\\hline
    
    \multirow{1}{*}{MobileBERT} & Prec. & {0.527} & {0.552} & {+4.74} & {0.602} & {0.493} & {-18.10} & {0.582} & {0.540} & {-7.21} & {0.648} & {0.558} & {-13.88}\\
    & Recall & {0.042} & {0.243} & {+478.57} & {0.462} & {0.720} & {+55.84} & {0.173} & {0.357} & {+106.35} & {0.449} & {0.662} & {+47.43}\\
    & F1 & {0.077} & {0.337} & {+333.74} & {0.522} & {0.585} & {+11.94} & {0.266} & {0.429} & {+61.15} & {0.530} & {0.605} & {+14.16}\\\hline
    
    \multirow{1}{*}{HyperHate-} & Prec. & {0.112} & {0.317} & {+183.0} & {0.364} & {0.474} & {+30.2} & {0.449} & {0.592} & {+31.8} & {0.760} & {0.607} & {-20.13}\\
    {Static}& Recall & {0.174} & {0.617} & {+254.6} & {0.442} & {0.639} & {+44.6} & {0.100} & {0.348} & {+248.0} & {0.080} & {0.740} & {+825.0}\\
    & F1 & {0.136} & {\textbf{0.418}} & {+207.3} & {0.399} & {0.544} & {+36.3} & {0.163} & {0.438} & {+167.97} & {0.144} & {\textbf{0.666}} & {+360.71}\\\hline
    
    \multirow{1}{*}{HyperHate-} & Prec. & {0.100} & {0.290} & {+190.00} & {0.377} & {0.538} & {+42.70} & {0.380} & {0.666} & {+75.26} & {0.386} & {0.600} & {+55.44}\\
    {Dynamic}& Recall & {0.153} & {0.540} & {+252.94} & {0.466} & {0.857} & {+83.9} & {0.158} & {0.479} & {+75.26} & {0.230} & {0.718} & {+212.174}\\
    & F1 & {0.120} & {0.377} & {+211.99} & {0.416} & {\textbf{0.661}} & {+55.59} & {0.223} & {0.557} & {+149.65} & {0.288} & {0.653} & {+126.79}\\\hline\hline
    
\end{tabular}}
\resizebox{\textwidth}{!}{
\begin{tabular}{ll|ccr|ccr|ccr|ccr}
     & & \multicolumn{3}{c}{SE-FN} & \multicolumn{3}{c}{WS-FN} & \multicolumn{3}{c}{DV-WS} & \multicolumn{3}{c}{WS-WH}\\
    Architecture & & Baseline & Augmented & (\%) & Baseline & Augmented & (\%) & Baseline & Augmented & (\%) & Baseline & Augmented &(\%)\\ \hline
    \multirow{1}{*}{CNN-GRU} & Prec. & {0.363} & {0.312} & {-14.04} & {0.092} & {0.167} & {+81.52} & {0.487} & {0.670} & {+37.57} & {0.299} & {0.347} & {+16.05}\\
    & Recall & {0.155} & {0.655} & {+322.58} & {0.274} & {0.320} & {+16.78} & {0.365} & {0.872} & {+138.90} & {0.609} & {0.817} & {+34.15}\\
    & F1 & {0.217} & {0.422} & {+94.56} & {0.137} & {0.219} & {+59.32} & {0.417} & {0.757} & {+81.6} & {0.401} & {0.487}& {+21.44}\\\hline
    
    \multirow{1}{*}{BERT} & Prec. & {0.445} & {0.257} & {-42.24} & {0.133} & {0.156} & {+17.29} & {0.718} & {0.668} & {-6.96} & {0.294} & {0.340}& {+15.64}\\
    & Recall & {0.169} & {0.500} & {+195.85} & {0.245} & {0.295} & {+20.40} & {0.712} & {0.861} & {+20.92} & {0.832} & {0.792}& {-4.80}\\
    & F1 & {0.244} & {0.339} & {+38.58} & {0.172} & {0.204} & {+18.37} & {0.714} & {0.752} & {+5.22} & {0.434} & {0.475}& {+9.50}\\\hline
    
    \multirow{1}{*}{CharBERT} & Prec. & {0.439} & {0.258} & {-41.23} & {0.135} & {0.158} & {
    17.03} & {0.721} & {0.659} & {-8.59} & {0.291} & {0.348} & {+19.58}\\
    & Recall & {0.162} & {0.503} & {+210.49} & {0.244} & {0.296} & {+21.31} & {0.715} & {0.852} & {+19.16} & {0.833} & {0.793} & {-4.80}\\
    & F1 & {0.236} & {0.341} & {+44.11} & {0.173} & {0.206} & {+18.52} & {0.717} & {0.743} & {+3.50} & {0.431} & {0.483} & {+12.14}\\\hline
    
    \multirow{1}{*}{RoBERTa} & Prec. & {0.502} & {0.281} & {-44.02} & {0.259} & {0.194} & {-25.09} & {0.717} & {0.741} & {+3.34} & {0.352} & {0.407} & {
    15.62}\\
    & Recall & {0.159} & {0.784} & {+393.08} & {0.144} & {0.54} & {+275.00} & {0.748} & {0.745} & {-0.40} & {0.534} & {0.574} & {+7.49}\\
    & F1 & {0.241} & {0.413} & {+71.30} & {0.185} & {\textbf{0.285}} & {+54.22} & {0.732} & {0.742} & {+1.47s} & {0.424} & {0.476} & {+12.25}\\\hline
    
    \multirow{1}{*}{Albert} & Prec. & {0.165} & {0.320} & {+93.93} & {0.280} & {0.145} & {-48.21} & {0.335} & {0.573} & {+71.04} & {0.351} & {0.313}& {-10.82}\\
    & Recall & {0.343} & {0.678} & {+97.66} & {0.120} & {0.271} & {+125.83} & {0.214} & {0.574} & {+168.22} & {0.406} & {0.705}& {+73.64}\\
    & F1 & {0.222} & {\textbf{0.434}} & {+95.13} & {0.168} & {0.188} & {+12.45} & {0.261} & {0.573} & {+119.59} & {0.376} & {0.433}& {+15.14}\\\hline
    
    \multirow{1}{*}{MobileBERT} & Prec. & {0.156} & {0.302} & {+93.58} & {0.276} & {0.249} & {-10.50} & {0.718} & {0.659} & {-8.21} & {0.329} & {0.379} & {+15.19}\\
    & Recall & {0.726} & {0.444} & {-38.84} & {0.129} & {0.201} & {+55.81} & {0.323} & {0.903} & {+179.56} & {0.763} & {0.691} & {-9.43}\\
    & F1 & {0.256} & {0.359} & {+39.97} & {0.175} & {0.221} & {+26.05} & {0.445} & {\textbf{0.761}} & {+71.00} & {0.459} & {\textbf{0.489}} & {+6.47}\\\hline
    
    \multirow{1}{*}{HyperHate-} & Prec. & {0.122} & {0.286} & {+134.42} & {0.134} & {0.181} & {+35.07} & {0.155} & {0.670} & {+332.25} & {0.276} & {0.288}& {+4.34}\\
    {Static}& Recall & {0.180} & {0.590} & {+227.77} & {0.115} & {0.352} & {+206.08} & {0.028} & {0.847} & {+2925.0} & {0.405} & {0.624}& {+54.07}\\
    & F1 & {0.145} & {0.385} & {+164.90} & {0.123} & {0.239} & {+93.14} & {0.047} & {0.748} & {+1477} & {0.328} & {0.394}& {+20.05}\\\hline
    
    \multirow{1}{*}{HyperHate-} & Prec. & {0.154} & {0.295} & {+91.55} & {0.124} & {0.163} & {+31.45} & {0.215} & {0.604} & {+180.93} & {0.258} & {0.298}& {+15.50}\\
    {Dynamic}& Recall & {0.293} & {0.822} & {+180.54} & {0.119} & {0.311} & {+161.34} & {0.097} & {0.636} & {+555.67} & {0.514} & {0.654}& {+27.23}\\
    & F1 & {0.201} & {\textbf{0.434}} & {+115.06} & {0.121} & {0.213} & {+76.11} & {0.133} & {0.619} & {+363.46} & {0.343} & {0.409}& {+19.17}\\\hline
\end{tabular}}
\end{table*}

%\end{thebibliography}
\end{document}
\endinput
%%
%% End of file `elsarticle-template-num.tex'.